\def\year{2021}\relax
\documentclass[letterpaper]{article}

\usepackage{authblk}

\usepackage{aaai21} 
\usepackage{times} 
\usepackage{helvet} 
\usepackage{courier} 
\usepackage[hyphens]{url} 
\usepackage{graphicx} 
\usepackage{graphicx} 
\usepackage{natbib}
\usepackage{caption}
\usepackage{amsmath}
\usepackage{amsfonts}
\usepackage{amssymb}
\usepackage{mathtools} 

\usepackage{balance}

\graphicspath{ {figs/} }
\usepackage{subcaption}

\usepackage{xcolor}

\usepackage{hyperref}
\hypersetup{
    colorlinks=true,
    linkcolor=black,
    filecolor=black,      
    urlcolor=black, 
    citecolor=black
}

\newcommand{\appendixlink}[1]{\href{https://arxiv.org/abs/2009.09961}{#1}}

\newcommand{\new}[1]{#1}

\usepackage{xspace}
\newcommand{\hbert}{SHERBERT\xspace}

\newcommand{\hide}[1]{}

\newcommand{\eg}{\textit{e.g.}\xspace}
\newcommand{\ie}{\textit{i.e.}\xspace}
\newcommand{\sect}{\S}

\newcommand\blfootnote[1]{%
  \begingroup
  \renewcommand\thefootnote{}\footnote{#1}%
  \addtocounter{footnote}{-1}%
  \endgroup
}

\newcommand{\papersite}{{\footnotesize \url{https://behavioral-data.github.io/CausalInferenceChallenges/}}}

\title{Adjusting for Confounders with Text: Challenges and an \\ Empirical Evaluation Framework for Causal Inference}

\author[1]{\Large \bfseries Galen Weld}
\author[1]{\Large \bfseries Peter West}
\author[2]{\Large \bfseries Maria Glenski}
\author[3]{\Large \bfseries David Arbour}
\author[3]{\Large \bfseries Ryan A. Rossi}
\author[1]{\Large \bfseries Tim Althoff}
\affil[1]{\normalsize University of Washington}
\affil[2]{\normalsize Pacific Northwest National Laboratory}
\affil[3]{\normalsize Adobe Research}

\begin{document}
\maketitle
\blfootnote{Galen Weld and Peter West contributed equally to this work.}

\begin{abstract}
Causal inference studies using textual social media data can provide actionable insights on human behavior.
Making accurate causal inferences with text requires controlling for confounding which could otherwise impart bias. Recently, many different methods for adjusting for confounders have been proposed, and we show that these existing methods disagree with one another on two datasets inspired by previous social media studies.
Evaluating causal methods is challenging, as ground truth counterfactuals are almost never available.
Presently, no empirical evaluation framework for causal methods using text exists, and as such, practitioners must select their methods without guidance.
We contribute the first such framework, which consists of five tasks drawn from real world studies. Our framework enables the evaluation of \textit{any} casual inference method using text.
\new{Across 648 experiments and two datasets,} we evaluate every commonly used causal inference method and identify their strengths and weaknesses to inform social media researchers seeking to use such methods, and guide future improvements.
We make all tasks, data, and models public to inform applications and encourage additional research.
\end{abstract}

\section{Introduction}\label{sec:intro}
The massive volume of social media data offers significant potential to help researchers better understand human behavior by making causal inferences. Researchers often formalize casual inference as the estimation of the
\textit{average treatment effect} ($ATE$) of a specific treatment variable (\eg therapy) on a specific outcome (\eg suicide)~\cite{rubin_2005_potential_outcomes, rosenbaum_2010, keith2020review}.
A major challenge is adjusting for \textit{confounders} (\eg comments mentioning depression) that affect both the treatment and outcome (depression affects both an individual's propensity to receive therapy and their risk of suicide)~\cite{keith2020review}.
Without adjusting for depression as a confounder, we might look at suicide rates among therapy patients and those not receiving therapy, and wrongly conclude that therapy causes suicide.

The gold standard for avoiding confounders is to assign treatment via a \textit{randomized controlled trial} (RCT).
Unfortunately, in many domains, assigning treatments in this manner is not feasible (\eg due to ethical or practical concerns).
Instead, researchers conduct \textit{observational studies}~\cite{rosenbaum_2010}, using alternate methods to adjust for confounders.

Text (\eg users' social media histories) can be used to adjust for confounding by training an NLP model to recognize confounders (or proxies for confounders) in the text, so that similar treated and untreated observations can be compared.
However, a recent review~\cite{keith2020review} finds that evaluating the performance of such methods is ``a difficult and open research question'' as \textit{true} $ATE$s are almost never known, and so, unlike in other NLP tasks, we cannot know the correct answer.
We find that this challenge is amplified, as methods disagree with one another on real world tasks (\sect\ref{sec:motivation}) -- how do we know which is correct?

\begin{figure}[t]
    \centering
    \includegraphics[width=.7\columnwidth]{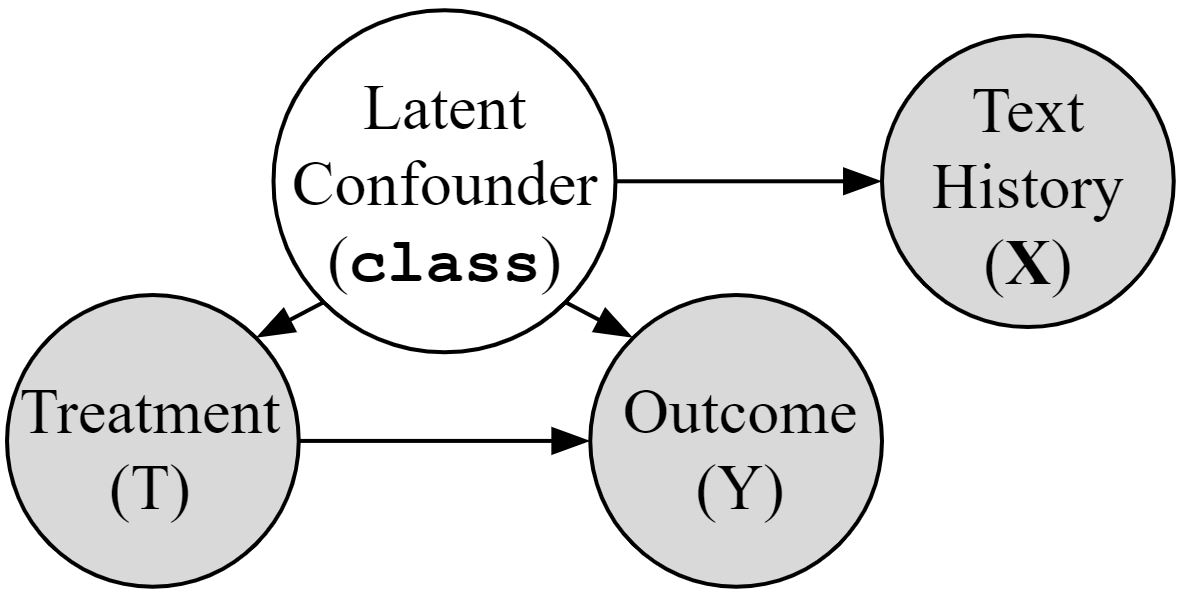}
    \caption{Causal graph representing the the context of our evaluation framework. All edges have known probabilities. While our framework naturally generalizes to more complex scenarios, we chose binary treatments (T) and outcomes (Y), and a binary latent confounder (\texttt{class}), as even in this simple scenario, current \new{methods} struggle.}
    \label{fig:causal_graph}
\end{figure}

Theoretical bounds on the performance of methods are almost never tight enough to be informative. We derive such bounds for the methods included here (\appendixlink{Appendix}~\ref{appendix:bounds}\footnote{\appendixlink{Link to Extended Paper with Appendix}}) and find that our empirical evaluation framework produces tighter bounds more than 99\% of the time.
As ground truth is almost never available, the only\footnote{With the extremely rare exception of \textit{constructed} observational studies, conducted with a parallel RCT.} practical method to evaluate causal inference methods is with semi-synthetic data, where synthetic treatments and/or outcomes are assigned to real observations, as in Fig.~\ref{fig:causal_graph} \cite{dorie2019acic, jensen2019comment, gentzel2019evaluating}.
While widely-used semi-synthetic benchmarks have produced positive results in the medical domain \cite{dorie2019acic}, no such benchmark exists for causal inference methods using text \cite{keith2020review}.

In this work, we contribute the first evaluation framework for causal inference with text (\sect\ref{sec:generation}).
Our framework is simple, principled, and can be applied to any method that produces an $ATE$ estimate given observed treatment assignments, outcomes, and text covariates. It includes a broad range of tasks challenging enough to identify where current methods fail and the most promising avenues for improvement. However, no single benchmark can adequately evaluate methods for \textit{every} application. As such, our framework can be easily extended to include additional tasks relevant to any application and potential confounder (\sect\ref{sec:generalizability}). 

Inspired by challenges from a wide range of studies \cite{johansson2016learning, dechoudhury2016mental_chi, choudhury2017language, falavarjani2017estimating, olteanu2017distilling, kiciman2018UsingLS, sridhar2018estimating, saha2019ASM, veitch2019embeddings, roberts2018adjusting}, our framework consists of five tasks (\sect\ref{sec:tasks}):
Linguistic Complexity, Signal Intensity, Strength of Selection Effect, Sample Size, and Placebo Test.
Each semi-synthetic task is generated from \new{public social media} users' profiles, perturbed with synthetic posts to create increasing levels of difficulty. \new{To increase the robustness of our approach, we evaluate methods on these tasks using real data from both Reddit and Twitter.} Not all of these tasks are exclusive to causal inference with text, yet all are important to a great deal of textual causal inference studies. As such, their evaluation in this context is important. Using these tasks, 
we evaluate the specific strengths and weakness of \new{9} widely-used methods and \new{3 common estimators, conducting 648 experiments.}

Concerningly, we find that almost every method predicts a false significant treatment effect when none is present, which could be greatly misleading to unwary practitioners (\sect\ref{sec:results}).
While we find that each method struggles with at least one challenge, methods leveraging recent, hierarchical, transformer-based architectures perform best, although such methods are not yet widely used \cite{keith2020review}.
These limitations and findings highlight the importance of continued research on the evaluation of causal inference methods for text.

The ICWSM community consists of researchers who work both on the development of casual inference methods, and practitioners who solve real world problems using causal inference.

\noindent
\textbf{For methods developers:} We make our framework publicly available\footnote{\label{papersite_footnote}\papersite} to enable the evaluation of \textit{any} causal inference method which uses text, and encourage the development of more robust methods. Our framework can be easily extended to include additional tasks.

\noindent
\textbf{For practitioners:} We identify strengths and weaknesses of commonly used methods, identifying those best suited for specific applications, and make these publicly available\footnotemark[3].

\section{Background and Related Work}\label{sec:related}

\subsubsection{Causal Inference with Social Media Data}
We formalize causal inference using notation from ~\citet{pearl1995causal}.
Given a series of $n$ observations (in our context, a social media user), each observation is a tuple $O_i = (Y_i, T_i, \mathbf{X}_i)$, where $Y_i$ is the outcome (\eg did user $i$ develop a suicidal ideation?), $T_i$ is the treatment (\eg did user $i$ receive therapy?), and $\mathbf{X}_i$ is the vector of observed covariates (\eg user $i$'s textual social media history).

The \textit{Fundamental Problem of Causal Inference} is that each user is either treated \textit{or} untreated, and so we can never observe \textit{both} outcomes.
Thus, we cannot compute the $ATE = \frac{1}{n} \sum^n_{i=1} Y_i\left[T_i = 1\right] - Y_i\left[T_i = 0\right]$ directly, and must estimate it by finding comparable treated and untreated observations.
To do so, it is common practice to use a model to estimate the \textbf{propensity score}, $\hat{p}(\mathbf{X}_i)\approx p(T_i=1|\mathbf{X}_i)$, for each observation $i$.
As treatments are typically known, propensity score models are effectively supervised classifiers, predicting $T_i$, given $\mathbf{X}_i$.
Matching, stratifying, or weighting using these propensity scores will produce an unbiased $ATE$ estimate if four assumptions hold: all confounders must be observed, outcomes for each user must not be affected by treatment assignments to other users (SUTVA), propensity scores must be accurate, and there must be overlap in the distribution of covariates in the treated and untreated groups (\textit{common support assumption}) \cite{rosenbaum_2010, hill2013common_support}. In practice, verifying these assumptions is difficult. In particular, ensuring that all confounding factors are observed is practically impossible in real world applications, hence the need for empirical evaluation.

\begin{figure*}[t]
    \centering
    \centerline{ \includegraphics[width=1.04\textwidth, trim={.5cm 0.12cm 1.71cm 0},clip]{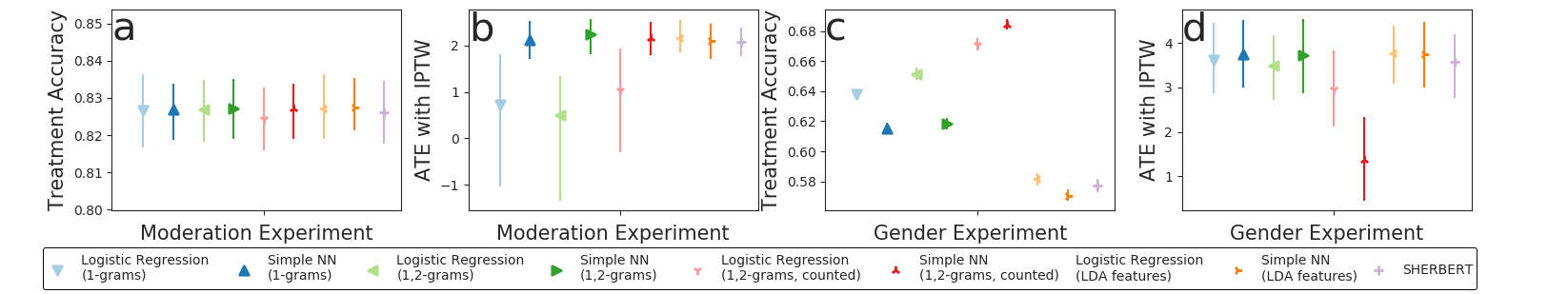} }
    \vspace{-1.2mm}
    \caption{Treatment accuracy and $ATE$ for both real world experiments, with bootstrapped 95\% confidence intervals. Note that for the Gender Experiment, the models with the highest accuracy have the lowest $ATE$.}
    \label{fig:real_world}
    \vspace{-2mm}
\end{figure*}

\subsubsection{Causal Inference and NLP}
Until recently, there has been little engagement between causal inference researchers and the NLP research community \cite{keith2020review}. 
There are many ways to consider text in a causal context, such as text as a mediator \cite{veitch2019embeddings,landeiro2016robust}, text as treatment \cite{wood_doughty2018challenges, egami2018make, fong2016discovery, tan2014effect, zhang2020quantifying}, text as outcome \cite{egami2018make, zhang2018conversations}, and causal discovery from text \cite{mani2000discovery, mirza2016catena}.
However, we narrow our focus to text \textit{as a confounder}.
This is an important area of research because the challenge of adjusting for confounding underlies most causal contexts, such as text as treatment or outcome \cite{keith2020review}.
Effective adjusting for confounding with text enables causal inference in any situation where observations can be represented with text -- \eg social media, news articles, and dialogue.

\subsubsection{Adjusting for Confounding with Text}
A recent review~\cite[Table~1]{keith2020review} summarizes common practices across a diverse range of studies.
Almost every method used in practice consists of two parts: a \textbf{propensity score model}, which uses some text representation to estimate propensity scores, and an \textbf{ATE estimator.} Since such propensity-score based methods are by far the most widely used, in this work, we focus on these methods.
While we do not evaluate non-propensity score methods such as doubly-robust methods~\cite{kang2007demystifying}, TMLE~\cite{schuler2017tmle}, and matching on values other than propensity scores \cite{roberts2018adjusting, mozer2020matching}, our framework's structure enables evaluation of \textit{any} $ATE$ estimation method that produces an $ATE$ estimate given observed treatment assignments, outcomes, and text covariates. We believe disentangling the challenges of widely-used methods is a key contribution before moving to more complex and less common methods. Evaluating these other methods is an important area of future work. 

Text representations used in propensity score models generally do not yet leverage recent breakthroughs in NLP, and roughly fall into three groups: those using uni- and bi-gram representations \cite{dechoudhury2016mental_chi, johansson2016learning, olteanu2017distilling}, those using LDA or topic modeling \cite{falavarjani2017estimating, roberts2018adjusting, sridhar2018estimating}, and those using neural word embeddings such as GLoVe \cite{pham2017deep}, fastText \cite{joulin-etal-2017-bag, chen2020causal}, or BERT \cite{veitch2019embeddings}, \cite{pryzant2018deconfounded}.
Three classes of estimators are commonly used to compute the $ATE$: \textbf{inverse probability of treatment weighting} (IPTW), propensity score \textbf{stratification}, and \textbf{matching}, either using propensity scores or, less frequently, some other distance metric.
In our evaluation, we separate the propensity score models from the ATE estimators to better understand each component's individual impact.

\subsubsection{Evaluation of Causal Inference}
In rare specialized cases, researchers can use the unbiased outcomes of a parallel RCT to evaluate those of an observational study, as in \citet{eckles2017bias}.
This practice is known as a \textit{constructed} observational study, and, while useful, is only possible where parallel RCTs can be conducted.
Outside these limited cases, proposed models are typically evaluated on synthetic data generated by their authors.
These synthetic datasets often favor the proposed model, and do not reflect the challenges faced by real applications \cite{keith2020review}.

Theoretical evaluation of causal inference methods is generally unhelpful, as theoretically derived performance bounds are mostly much less informative than those derived empirically~\cite{arbour2019permutation}. In this work, we compute the theoretical bounds and find that they are so loose as to not effectively guide practitioners in selecting methods. Our empirical evaluation framework based on realistic tasks produces tighter bounds in more than 99\% of cases (\appendixlink{Appendix}~\ref{appendix:bounds}).

Outside of the text domain, \textit{widely used} empirical evaluation datasets have been successful, most notably the 2016 Atlantic Causal Inference Competition \cite{dorie2019acic}, and a strong case has been made for the empirical evaluation of causal inference models \cite{gentzel2019evaluating, jensen2019comment, lin2019universal}.
In the text domain, matching approaches have been evaluated empirically \cite{mozer2020matching}, but this approach evaluates only the quality of \textit{matches}, not the causal effect estimates.
In contrast, our work applies to all estimators, not just matching, and evaluates the entire causal inference pipeline.

\section{Current Models Disagree}\label{sec:motivation}
Recent causal inference papers \cite{veitch2019embeddings, roberts2018adjusting, dechoudhury2016mental_chi, chandrasekharan2017ban, bhattacharya2016information} have used social media histories to adjust for confounding.
Each of these papers uses a different \new{propensity score} model: BERT in \citet{veitch2019embeddings}, topic modeling in \citet{roberts2018adjusting}, logistic regression in \citet{dechoudhury2016mental_chi}, \new{Mahalanobis distance matching in \citet{chandrasekharan2017ban}, and Granger Causality in \citet{bhattacharya2016information}.}
For all of these studies, ground truth causal effects are unavailable, and so we cannot tell if the chosen model was correct.
However, we \textit{can} compute the accuracy of their propensity scores (accuracy of a binary classifier predicting treatment assignment), and see if their $ATE$ estimates agree---if they don't, then at most one disagreeing model can be correct.

\subsubsection{Methods}
We conducted two experiments using real world data from Reddit, inspired by these recent papers.
In the \textbf{Moderation Experiment}, we test if having a post removed by a moderator impacts the amount a user later posts to the same community again. 
In the \textbf{Gender Experiment}, we use data from \citet{veitch2019embeddings} to study the impact of the author's gender on the score of their posts.
For details on data collection, see \appendixlink{Appendix}~\ref{appendix:motivation_experiments}.

\subsubsection{Results}
Comparing the performance of nine different \new{methods} (Fig.~\ref{fig:real_world}), we find that all models have similar treatment accuracy in the Moderation Experiment. However, the models using 1,2-gram features perform better in the Gender Experiment than the LDA and \hbert models.
Most importantly, we see that while many confidence intervals\footnote{In these experiments, as well as all following experiments, results are reported with bootstrapped 95\% confidence intervals computed by resampling from the population and recomputing the estimators (\sect\ref{sec:estimators}) and evaluation metrics. } overlap, there are notable differences between ATE estimates for different models, even when treatment accuracy is nearly identical (Fig.~\ref{fig:real_world}a,b). That some ATE estimates' confidence intervals overlap 0 while others do not (Fig.~\ref{fig:real_world}b) indicates that some models find nonzero treatment effects at the common $p=0.05$ threshold while others do not. Lastly, we note that models with the highest treatment accuracy tend to have the lowest ATE estimates (Fig.~\ref{fig:real_world}c,d).

\subsubsection{Implications}
This should come as a great concern to the computational social science research community.
We do not know which model may be correct, and we do not know whether there may be a more accurate model that would even further decrease the estimated treatment effect. 
We derive theoretical bounds and compute them (\appendixlink{Appendix}~\ref{appendix:bounds}), finding that in 99+\% of cases, these bounds are looser than those computed empirically using our framework, making them less useful for model selection.
This concern underlines the importance and urgency of empirical evaluation for causal inference with text, and motivates our contribution.
Next, we describe key challenges in adjusting for confounding with text and present a principled evaluation framework that highlights these challenges and generates actionable insights for future research.

\section{Challenges for Causal Inference with Text}\label{sec:complexity}
Using the common setting of real social media histories~\cite{dechoudhury2016mental_chi, olteanu2017distilling, veitch2019embeddings, choudhury2017language, falavarjani2017estimating, kiciman2018UsingLS, saha2019ASM, roberts2018adjusting}, we identify five challenges consistently present when representing natural language for causal inference:
\begin{enumerate}
\setlength\topsep{-3pt}
\setlength\itemsep{-3pt}
    \item \textbf{Linguistic Complexity:}
    Natural language uses a diverse set of tokens to express related underlying meaning.
    Someone who struggles with mental health might write ``I feel depressed'' or ``I am isolated from my peers,'' which have distinct tokens but both may be indicative of depression.
    \textit{Can models recognize a range of expressions which are correlated with treatment?}
    \item \textbf{Signal Intensity:} Some users only have a few posts that contain a specific signal (such as poor mental health) whereas others may have many posts with this signal. Signals are especially weak when posts containing the signal constitute only a small fraction of a user's posts. {\it Can models detect weak signals?}
    \item \textbf{Strength of Selection Effect:} Many studies have few comparable treated and untreated users \cite{li2018addressing, crump2009dealing}. {\it Can models adjust for strong selection effects? }

    \item \textbf{Sample Size:} Observational studies often face data collection limitations.\footnote{In \citet[Table~1]{keith2020review}, 8/12 studies had fewer than 5,000 observations, and 4/12 had fewer than 1,000.}
    {\it Can models perform well with limited data samples?}
    \item \textbf{Placebo Test:} Oftentimes, no causal effect is present between a given treatment and an outcome. {\it Do models falsely predict causality when none is present?}
\end{enumerate}

While natural language is far more complex than any finite set of challenges can capture, the five we have chosen to highlight are challenges that regularly need to be addressed in many causal inference applications that use natural language.
This set of challenges was developed by reviewing a broad set of existing studies \cite{johansson2016learning, dechoudhury2016mental_chi, choudhury2017language, falavarjani2017estimating, olteanu2017distilling, kiciman2018UsingLS, sridhar2018estimating, saha2019ASM, veitch2019embeddings, roberts2018adjusting} and identifying commonalities.
While the strength of selection effect, sample size, and placebo test challenges are not \textit{exclusive} to causal inference with text, these challenges are present in most real world studies, and as such, a holistic evaluation framework must consider them.
These five challenges also cover three key concepts of model performance: generalizability (\textit{linguistic complexity}), sensitivity (\textit{signal intensity}, \textit{strength of selection effect}), and feasibility (\textit{sample size}, \textit{placebo test}) that are critical for comprehensive evaluation.
To produce our evaluation framework, we derive a concrete task from each challenge.

\section{Framework for Evaluation}\label{sec:generation}  
We generate five tasks, each with discrete levels of difficulty, and corresponding semi-synthetic task datasets based on real social media histories. 
Without the semi-synthetic component, it would not be possible to empirically evaluate a model, as we would not know the true $ATE$ or propensity scores.
By basing our user histories on real data, we are able to include much of the realism of unstructured text found `in the wild.'
This semi-synthetic approach to evaluation preserves the best of both worlds: the empiricism of synthetic data with the realism of natural data \cite{jensen2019comment, gentzel2019evaluating, jensen2019comment}.
 
 \begin{figure}[t]
    \centering
    \includegraphics[width=.7\columnwidth]{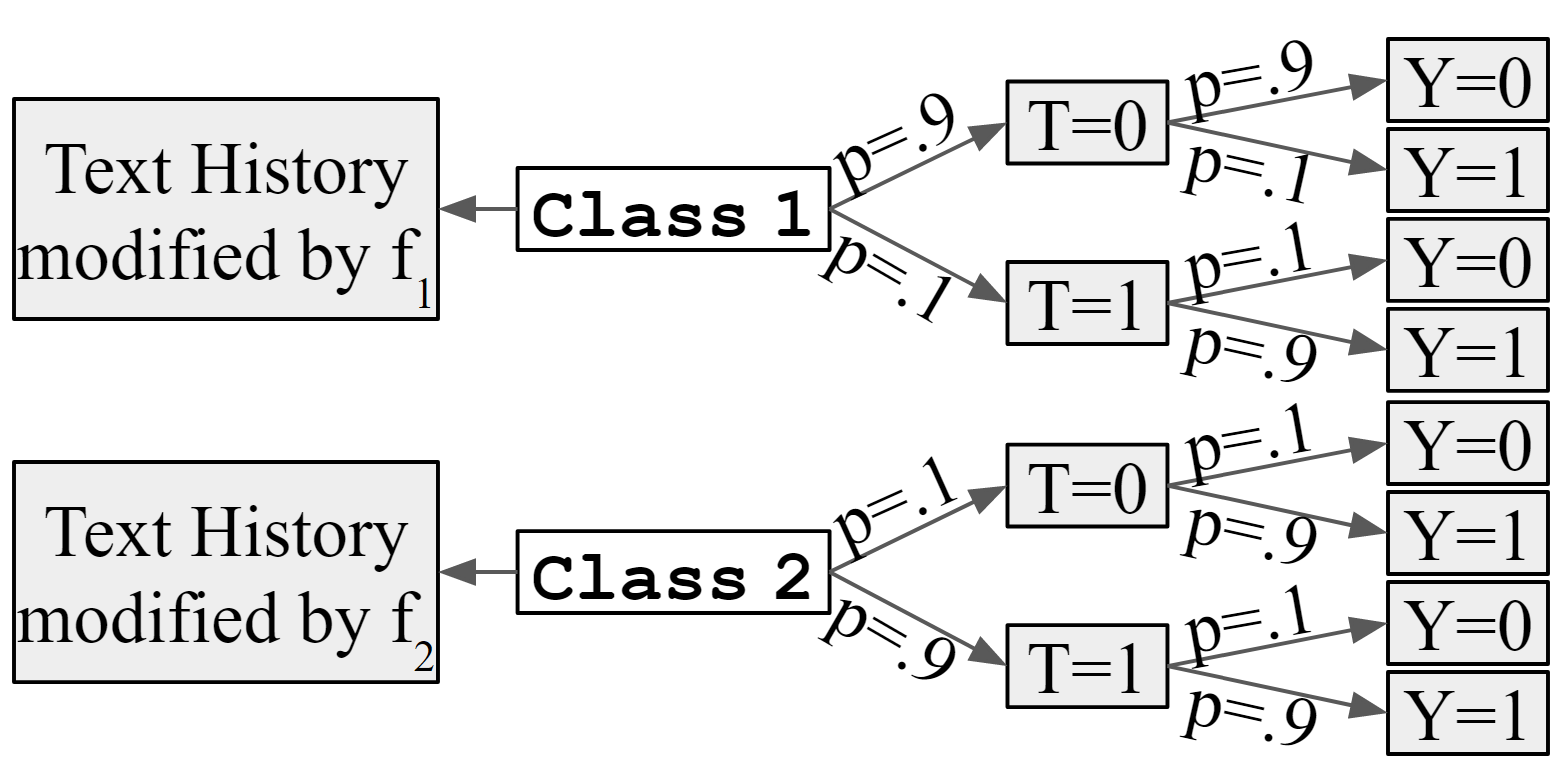}
    \caption{Users are first randomly divided into two latent (unobserved) classes with a 50/50 split, and their text histories have synthetic posts inserted specific to each task. Observed binary treatments and outcomes are assigned with conditional probabilities such that \texttt{Class 1} has a true $ATE$ of .8, and \texttt{Class 2} has a true $ATE$ of 0. Since the classes are balanced, the overall true $ATE$ is .4.
    }
    \label{fig:assignment}
\end{figure}

\subsection{Semi-Synthetic Dataset Generation} \label{sec:semisynthetic_data}
While the method for generating a semi-synthetic dataset can be arbitrarily complex, we seek the simplest approach which is able to identify and explain where existing methods fail. We generate our datasets according to a simplified model of the universe; where all confounding is present in the text, and where there are only two types of people, \texttt{class 1} and \texttt{class 2} (Fig.~\ref{fig:causal_graph}).
In the context of mental health, for example, these two classes could simply be people who struggle with depression (\texttt{class 1}), and those who don't (\texttt{class 2}).
If models struggle on even this simple two-class universe, as we find, then it is highly unlikely they will perform better in the more complex real world.
In this universe, the user's (latent) class determines the probability of treatment and outcome conditioned on treatment.
Dependent on class, but independent of treatment and outcome is the user's comment history, which contains both synthetic and real posts that are input to the model to produce propensity scores. As such, the comment history is an observed proxy for the \texttt{class} confounder.

We produce each dataset using a generative process, as shown in Fig.~\ref{fig:assignment}.
For each task, we start with the same collection of real world user histories from public Reddit \new{or Twitter} profiles.
We randomly assign an equal number of users to \texttt{class 1} and \texttt{class 2}.
Into each profile, we insert synthetic posts using a function $f_n$ for $\texttt{class }n$ specific to each task, described in \sect\ref{sec:tasks}.
We assign binary treatments (conditioned on class) and binary outcomes (conditioned on class and treatment) according to a known probability distribution (Fig.~\ref{fig:assignment}).
These outcomes and treatments could represent anything of interest, and they need not be binary.

To estimate the $ATE$, there must be overlap between the treated and untreated groups (\textit{common support}), so
we cannot make all users in \texttt{class 1} treated and all users in \texttt{class 2} untreated. Instead, users in \texttt{class 1} are predominantly but not always assigned to treatment (with a .9/.1 split), and users in \texttt{class 2} are predominantly but not always assigned to control (also with a .9/.1 split), in order to ensure overlap of covariates between the treated and control groups.
This overlap provides common support, and thus our observations do not necessitate trimming \cite{crump2009dealing, lee2011weight, yang2018asymptotic}. 

Once a treatment has been assigned according to the class' probabilities, a positive outcome is assigned with probability $.9$ (treated) and $.1$ (untreated) for \texttt{class 1}, and $.9$ regardless of treatment for \texttt{class 2}.
Thus, \texttt{class 1} has a true $ATE$ of .8, and \texttt{class 2} has a true $ATE$ of 0. Since the the two classes are balanced, the overall true $ATE$ is .4.
The objective for propensity score models is to recover the treatment probabilities for each class, which are then used to estimate the true $ATE$.

\subsubsection{Real World User Histories}
\label{sec:real_data}
\new{In order to maximize generalizability, we experiment with both Reddit and Twitter user histories} as the real world component of our semi-synthetic datasets.
\new{Reddit and Twitter are natural data sources as they are both publicly accessible, and widely used for relevant research} \cite{medvedev2019anatomy, Yu2020ABO}.

We downloaded all Reddit comments for the 2014 and 2015 calendar years from the Pushshift archives \cite{baumgartner2020pushshift} and grouped comments by user. After filtering out users with fewer than 10 comments, we randomly sampled 8,000 users and truncated users' histories to a maximum length of 60 posts for computational practicality.\footnote{The resulting set of \new{Reddit} users had a mean of 41.07 posts per user, mean of 37.37 tokens per post, and a mean of 1523.28 tokens per user.}
These users were randomly partitioned into three sets: a 3,200 user training set, an 800 user validation set, and a 4,000 user test set used to compute Treatment Accuracy and $ATE$ Bias.

\new{
To gather Twitter data, we used a similar method as that used for Reddit. We used the Streaming API to produce a random sample of all public tweets posted in December 2020, then randomly sampled users from this sample and used the Twitter API to gather their complete Tweet histories. As with Reddit histories, we filtered out users with fewer than 10 Tweets, and truncated the histories of users with more than 60 Tweets, then randomly partitioned the resulting users into a training set, a validation set, and a test set, each of the same size as their Reddit data counterparts.\footnote{\new{The resulting set of Twitter users had a mean of 57.76 posts per user, mean of 19.90 tokens per post, and a mean of 1149.59 tokens per user.}}
}

\subsubsection{Synthetic Posts}\label{sec:synthetic_data}
When generating semi-synthetic tasks, we insert three types of synthetic posts, representative of major life events that could impact mental health, into real users' \new{Reddit and Twitter} histories. Examples are given here, and are listed completely in \appendixlink{Appendix}~\ref{appendix:templates}:
\begin{itemize}
\setlength\topsep{-2pt}
\setlength\itemsep{-3pt}
\item {\it Sickness Posts} describe being ill (\eg `The doctor told me I have AIDS', `How do I tell my parents I have leukemia?').
We vary both the illness, as well as way the it is expressed.
\item {\it Social Isolation Posts} indicate a sense of isolation or exclusion.
(`I feel so alone, my last friend said they needed to stop seeing me.', `My wife just left me.')
\item {\it Death Posts} describe the death of companion (\eg `I just found out my Mom died', `I am in shock. My son is gone.'). We vary the phrasing as well as the companion.
\end{itemize} 
\vspace{-0.5\baselineskip}
While these synthetic posts are drawn from the mental health domain, which is commonly represented among previous studies~\cite{dechoudhury2016mental_chi, choudhury2017language, saha2019ASM}, the applicability of our framework is not limited to this specific context. These synthetic posts test a model's ability to recognize different tokens, and the specific tokens used are less critical. Furthermore, the concrete choice of mental health language has the benefit of making the user histories human-readable. Our framework can be easily extended by modifying the synthetic posts to include tokens from different domains (\sect\ref{sec:generalizability}).
\subsection{Tasks}\label{sec:tasks}
We consider five tasks focused around the previously described common challenges for text-based causal inference methods.
\subsubsection{Linguistic Complexity}
\label{task:ling_comp}
This task tests a model's ability to recognize a diverse set of tokens as being correlated with treatment. 
We increase the difficulty in four steps by increasing the diversity of synthetic sentences inserted into user histories assigned to \texttt{class 1} (\ie the linguistic complexity of the dataset):
$f_1$ initially appends the same Sickness Post to the end of each \texttt{class 1} user's history;
At the second level of difficulty, $f_1$ selects a Sickness Post uniformly at random;
At the third level, $f_1$ selects either a Sickness or Social Isolation Post; and at the fourth level, $f_1$ selects a Sickness, Social Isolation, or Death Post.
For each level of difficulty, $f_2$ is the identity function, \ie user histories assigned to \texttt{class 2} are unchanged.

\subsubsection{Signal Intensity}
\label{task:signal}
This task tests a model's ability to distinguish between the number of similar posts in a history.
There are two levels of difficulty.
At the easier level, $f_1$ appends 10 randomly sampled (with replacement) Sickness Posts, while $f_2$ is the identity function.
At the harder level, $f_1$ appends only three Sickness Posts, while $f_2$ appends one.

\subsubsection{Strength of Selection Effect}
\label{task:selection_effect}
In this and the following tasks, we do not vary $f_1$ or $f_2$.
For \textit{Strength of Selection Effect}, we make causal inference more challenging by increasing the strength of the selection effect, decreasing the overlap between treated and untreated users.
We test two levels of difficulty: a weaker selection effect (easier) with the same .9/.1 split to assign the majority of \texttt{class 1} to the treated group and \texttt{class 2} to the control group.
For the stronger selection effect (harder), we modify the generation framework to increase this split for \texttt{class 1} to .95/.05.
For both the weak and strong selection effects, we use $f_1$ to append a single random Sickness Post and $f_2$ as the identity function.
Outcome probabilities, conditioned on treatment, are identical to previous tasks.


\subsubsection{Sample Size}
\label{task:amount_data}
In this task, we test how the models' performance drops off as the amount of available training data is reduced.\footnote{In \citet[Table~1]{keith2020review}, 8/12 studies had fewer than 5,000 observations, and 4/12 had fewer than 1,000.}
As before, we use $f_1$ to append a single random Sickness Post and $f_2$ as the identity function.
For the easiest case, we train on all 3,200 users' histories in the training set.
We then create smaller training sets by randomly sampling subsets with 1,600 and 800 users.

\subsubsection{Placebo Test}
\label{task:null}
The final task assesses a model's tendency to predict a treatment effect when none is present.
To do so, we must have asymmetric treatment probabilities between \texttt{class 1} and \texttt{class 2}. Without this asymmetry, the unadjusted estimate would be equal to the true $ATE$ of zero.
We use the same asymmetric \texttt{class 1} treatment split as in the Strength of Selection Effect task.

We set $P(Y=1|T=0,\texttt{class=1})=.05$, $P(Y=1|T=1,\texttt{class=2})=.95$, and the opposite for $Y=0$.
This gives a treatment effect of +.9 to \texttt{class~1} and a treatment effect of -.9 to \texttt{class~2}, making the true $ATE$ for the entire task equal 0.
As in previous tasks, $f_1$ appends one random Sickness Post and $f_2$ is the identity function.

\new{
A potential limitation of these tasks may be the placement of synthetic posts at the end of histories, or differences in the length of histories. However in \sect\ref{sec:order_num_posts} we show that our results suggest that these potential limitations are very unlikely to affect the validity of evaluation. Futhermore, the framework may be extended (\sect\ref{sec:generalizability}) to further evaluate methods on these aspects.}
\subsection{Generalizability of Evaluation Framework}\label{sec:generalizability}
\new{A key tenet of our evaluation framework is its extensibility. Here, we provide a summary of how the framework can be extended, with step by step instructions for addition of continuous outcome values and new tasks. Additional details and resources are available on our website.\footnote{\papersite}}

\subsubsection{Non-binary Treatments and Outcomes}
The evaluation framework presented here tests methods' ability to handle five distinct challenges which are relevant to many real world studies. It can be applied to \textit{any} causal inference method which uses text to adjust for confounding. As our framework is the first such framework, we focus on the simplest and most broadly applicable cases: binary treatments and outcomes, as these are the most commonly used in practice. Out of 14 recent casual inference studies, 13 used binary treatments~\cite{keith2020review}. While the methods evaluated here have been generalized to handle observations with continuous treatment values (\ie, dose-response observations)~\cite{hirano2004continuous}, these methods are rarely used in practice---we are not aware of a single such application which uses text data. \new{However,} our framework trivially generalizes to cases with continuous outcome values, and can be easily modified to include continuous treatments and multiple confounders, including unobserved confounders. 

\new{To modify the framework to use continuous outcomes:}
\begin{enumerate}
    \item \new{Select a random distribution to use to assign outcome values, conditioned on treatment. A straightforward option would be to use normal distributions with $\sigma^2=0.3$ and mean = .1 for (\texttt{class 1}, $T=0$) and mean = .9 for (\texttt{class 1}, $T=1$), (\texttt{class 2}, $T=0$), and (\texttt{class 2}, $T=1$). Any distribution may be used that provides reasonable common support as well as sufficient confusion and selection effects.}
    \item \new{Train propensity score models as with binary outcomes.}
    \item \new{Compute ATE estimates using IPTW, matching or stratification estimators, all of which can be applied to continuous outcomes without modification.}
    \item \new{Compute bias by taking the difference from the true ATE, which can be derived from the known treatment and outcome probability distributions.}
\end{enumerate}

\noindent
\new{Steps 2-4 are identical to the existing framework, as previously described.}

\subsubsection{Types of Confounding}
It is difficult or impossible to know exactly what types of confounders threaten the accuracy of real world causal inferences, and even simply making a reasonable guess requires substantial domain expertise~\cite{rosenbaum_2010}. As such, our evaluation framework simulates confounders which are common to many topics of interest to the ICWSM community, where specific `indicator' passages or posts in a longer text document are correlated with a treatment and/or outcome of interest. These `indicator' confounders have been shown to be common in the mental health~\cite{choudhury2014self_disclosure}, personal identity~\cite{haimson2018transgender_disclosure}, and fake news~\cite{talwar2019fake_news} domains. While our five tasks focus on these common types of confounder, tasks can be easily added to cover other types of confounders, for example by adding a task which applies some rewriting transformation (\eg desirability~\cite{wang2019lexical_choice, pryzant2018deconfounded} or gender~\cite{wang2019lexical_choice}) to the histories of \texttt{class 1} and/or \texttt{class 2}.

\new{To add an additional task which applies a rewriting transformation:}
\begin{enumerate}
    \item \new{Write a new $f_1$ function which, for example, replaces all occurrences of the token `happy' with `sad.'}
    \item \new{Write a new $f_2$ function which, for example, replaces all occurrences of the token `sad' with `happy.'}
    \item \new{Use the same treatment and outcome probabilities as used for other tasks, or, optionally, modify these probabilities to decrease or increase the overlap between the treated and untreated groups, as in the Strength of Selection Effect task.}
    \item \new{Train and evaluate methods' treatment accuracy and bias of the ATE using the known true ATE.}
\end{enumerate}
\section{Causal Inference Pipeline}\label{sec:models}
So many methods for adjusting for confounding with text have been proposed recently that is not possible to evaluate every one in a single paper. Instead, we focus on the most commonly used methods, those which are based upon propensity scores.
A recent review found that propensity score-based methods are used in 12/13 recent studies~\cite[Table~1]{keith2020review}.
Evaluating less commonly used methods (such as doubly robust methods~\cite{mayer2019doubly}) is an important area of future work, and our evaluation framework can be applied to \textit{any} method for adjusting for confounding with text.
The methods we evaluate here consist of three parts: a text representation, a propensity score model, and an $ATE$ estimator.

\subsection{Text Representations \& Propensity Score Models}\label{sec:ps_models}
The \textbf{Oracle} uses the true propensity scores, which are known 
in our semi-synthetic evaluation framework (Fig.~\ref{fig:assignment}).
The Oracle provides an upper-bound on model performance, only differing from the theoretical optimum due to finite sample effects.

We include an \textbf{Unadjusted Estimator}, which uses the naive method of not adjusting for selection effects, producing an estimated treatment effect of $\bar{Y}_{T=1}-\bar{Y}_{T=0}$, and as such is a lower-bound for models that attempt to correct for selection effects.

We train a \textbf{Simple Neural Net} (with one fully connected hidden layer) in four variants with different text representations: 1-grams with a binary encoding, 1,2-grams with a binary encoding, 1,2-grams with counts, and Latent Dirichlet Allocation (LDA) features \cite{Blei2003LatentDA} based on 1,2-grams, counted.
We also train \textbf{Logistic Regression} models on the same four text representations.
Vocabulary sizes for n-gram methods are included in \appendixlink{Appendix}~\ref{appendix:vocab_size}.


Finally, we propose and evaluate a novel cauSal HiERarchical variant of BERT, which we call {\bf \hbert}. 
\hbert expands upon Causal BERT proposed by \citet{veitch2019embeddings}, which is too computationally intensive to scale to user histories containing more than 250 tokens, let alone ones orders of magnitude longer, such as in our tasks.
In \hbert, we use one pretrained BERT model per post to produce a post-embedding (\appendixlink{Appendix}~\ref{appendix:sherbert_architecture} Fig.~\ref{fig:herbert}), followed by two hierarchical attention layers to produce a single embedding for the entire history, with a final linear layer to estimate the propensity score.
This architecture is similar to HIBERT~\cite{zhang-etal-2019-hibert}, but is faster to train on long textual histories, as \hbert fixes the pretrained BERT components. More details on \hbert are given in \appendixlink{Appendix}~\ref{appendix:model_implementation}.

\subsection{Average Treatment Effect Estimators}\label{sec:estimators}
We consider three commonly used $ATE$ estimators -- IPTW, stratification, and matching. All three estimators use propensity scores but differ in how they weight or group relevant samples.

\textbf{Inverse Propensity of Treatment Weighting} estimates the $ATE$ by weighting each user by their relevance to selection effects:
\vspace{-0.5\baselineskip}
  \begin{align*} 
    \widehat{ATE}_{\texttt{IPTW}} = \sum_{i=1}^n \frac{ ( 2*T_i - 1)*Y_i }{\hat{p}_{T_i} (\mathbf{X}_i) * \left[\sum_{j=1}^n \frac{1}{\hat{p}_{T_j} (\mathbf{X}_j)}\right] } 
\end{align*} 
\vspace{-0.75\baselineskip}

\noindent
where $T_i$, $Y_i$, and $X_i$ are treatment, outcome, and features for sample $i$, and $\hat{p}_T{(\mathbf{X})}$ is the estimated propensity for treatment $T$ on features $\mathbf{X}$. Use of the Hajek estimator~\shortcite{hajek1970} adjustment improves stability compared to simple inverse propensity. 

\textbf{Stratification} divides users into strata based on their propensity score, and the $ATE$ for each is averaged:
$
    \widehat{ATE}_{\texttt{strat}} = \frac{1}{n}\sum_k n_k*\widehat{ATE}_k
$

\noindent
where $n$ is the total number of users, $n_k$ is the number of users in the k-th stratum, and $\widehat{ATE}_k$ is the unadjusted $ATE$ within the k-th stratum. We report results on 10 strata divided evenly by percentile, but results are qualitatively similar for other $k$.

\textbf{Matching} can be considered as a special case of stratification, where each strata contains only one treated user.
We find that matching produces extremely similar results to stratification, and therefore we include details of our \new{matching} approach and results in \appendixlink{Appendix}~\ref{appendix:matching}.

\subsection{Metrics for Evaluation}\label{sec:metrics}
Our semi-synthetic tasks are generated such that we know the true $ATE$ and thus can compute the {\bf Bias of $\widehat{{\bf ATE}}$}.
A bias of zero is optimal, indicating a correct estimated $ATE$.
The greater the bias, positive or negative, the worse the model performance.
This is the primary metric we use in evaluation, and we compute it for both $\widehat{ATE}_{\texttt{strat}}$ and $ \widehat{ATE}_{\texttt{IPTW}}$.
We also consider {\bf Treatment Accuracy}, the accuracy of the propensity score model's predictions of binary treatment assignment.
While higher accuracy is often better, high accuracy does not guarantee low bias and often is instead indicative of strong selection effects.
Furthermore, we include two additional metrics in \appendixlink{Appendix}~\ref{appendix:additional_metrics}: First, the \textbf{Mean Squared Error of IPTW weights}, which captures the calibration of propensity scores probabilities and resulting weights, and second, the \textbf{Spearman's Rank Correlation.} In some cases, even if \textit{absolute} propensity scores are incorrect, their \textit{relative} rank may still contain useful information that could be exploited in stratification-based methods. The Spearman Correlation measures the correlation between the true and estimated propensity scores for each model, with a value of 1 indicating a perfect ranking.

\begin{figure*}
    \centering
    \centerline{\includegraphics[height=.94\textheight]{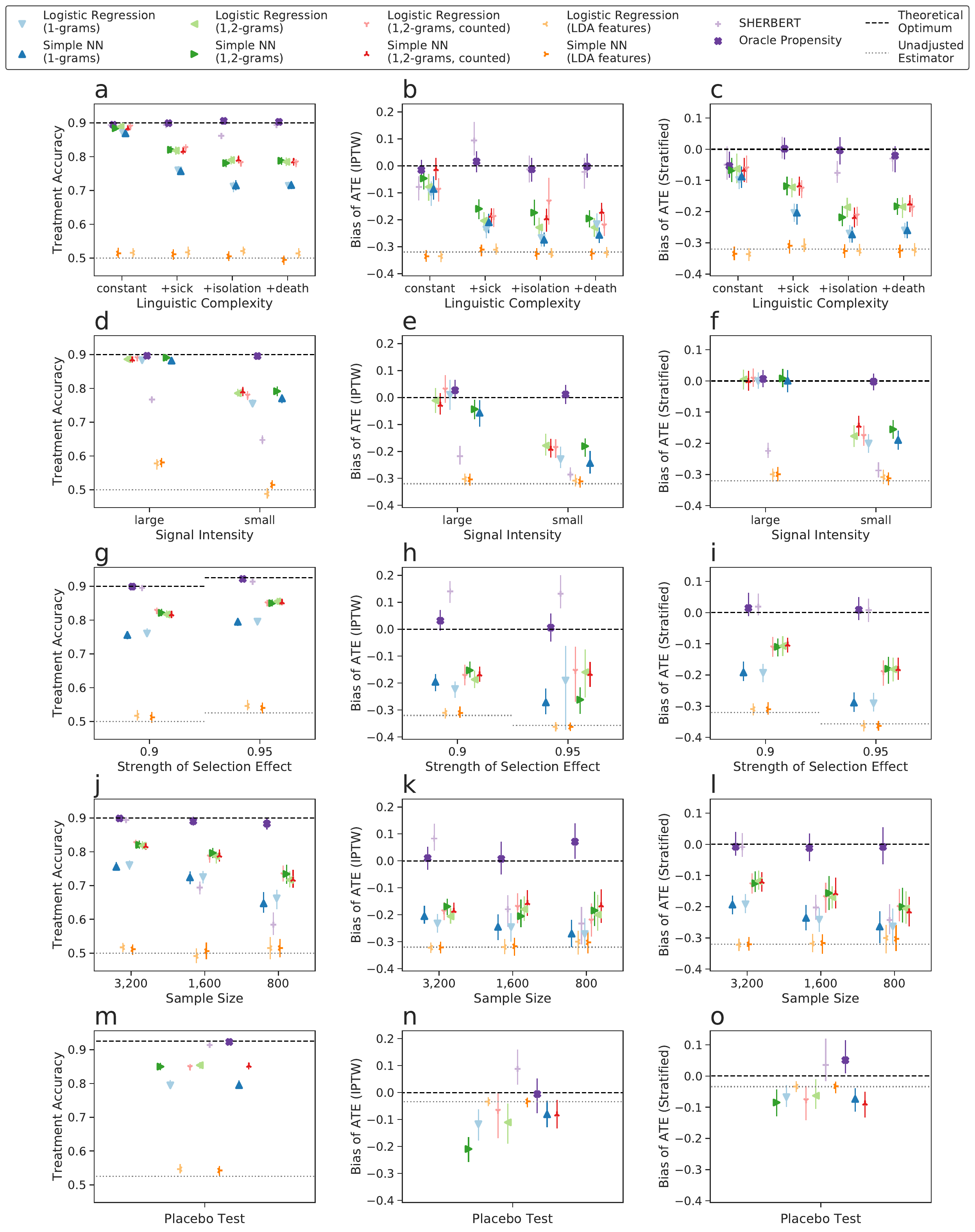}}
    \caption{Results for tasks \new{computed with Reddit data}, with bootstrapped 95\% confidence intervals, perturbed along the x-axis for readability. Columns represent metrics, and rows correspond to tasks.
    Within each plot, difficulty increases from left to right. 
    \hbert generally does well, especially on \textit{Strength of Selection Effect} and \textit{Placebo Test}, but struggles on \textit{Signal Intensity}. 
    }
    \label{fig:results}
\end{figure*}

\section{Results of Evaluation}\label{sec:results} 
\new{
We apply all five tasks of our evaluation framework to both Reddit and Twitter social media data. We found virtually identical results between the two datasets (\eg Treatment Accuracies within 1.9\% of one another), and so for brevity we report primarily on Reddit results here. Complete Twitter results and discussion are included in \appendixlink{Appendix}~\ref{appendix:twitter_results}.
}

\subsubsection{Transformers better model relevant linguistic variation} 
The Linguistic Complexity task shows many trends in the results manifest in other tasks,
including treatment accuracy clustering by text representation (Fig.~\ref{fig:results}a).
\hbert performs well, with uni- and bi-gram \new{methods} somewhere in between. Accuracy correlates fairly well with bias (Fig.~\ref{fig:results}b,c).
As in nearly all tasks, LDA \new{methods} perform worst, not even outperforming the unadjusted estimator. This is likely because LDA uses an unsupervised (agnostic to treatment) method to generate a compressed feature set that is likely to miss key features when they comprise only a small part of the overall user history.

\subsubsection{Transformer models struggle with counting and ordering} The Signal Intensity task requires \new{methods} to effectively `count' the number of posts to distinguish between classes.'
Here, n-gram \new{methods} outperform \hbert (Fig.~\ref{fig:results}e,f) This suggests that order embeddings are an important inclusion for future transformer-based \new{methods}. LDA \new{methods} perform slightly better than unadjusted, due to the stronger presence of tokens correlated with treatment.

\subsubsection{High accuracy often reflects strong selection effects, not low \textit{ATE} bias} In the Strength of Selection Effect task, we decrease the overlap in propensity scores between treated and untreated users, which makes it \textit{easier} to distinguish between the two groups. We see corresponding \textit{increases} in Treatment Accuracy (Fig.~\ref{fig:results}g), 
however, bias worsens (Fig.~\ref{fig:results}h,i). In \appendixlink{Appendix}~\ref{appendix:additional_metrics}, we also consider the Spearman Correlation, which evaluates propensity scores not on their \textit{absolute} accuracy, but on their \textit{relative} ranking, as in theory it is possible for inaccurate propensity scores to carry useful information. We expect the Stratified and Matching Estimators to perform better than IPTW in these cases, \eg in the Signal Intensity Task, where the stratified estimate has lower bias than IPTW (Fig.~\ref{fig:results}e,f) and the Spearman Correlation is higher than Treatment Accuracy at the harder difficulty level (\appendixlink{Appendix}~\ref{appendix:additional_metrics} Fig.~\ref{fig:additional_metrics}d,f).

In context of observational studies, \new{methods} with high treatment accuracy should be used with extreme caution --- high accuracy likely reflects treated and control groups that are too disjoint for any meaningful comparison to be drawn. In this case, the common support assumption is violated, preventing causal inference.
This highlights the importance of empirical evaluation of the \textit{complete} causal inference pipeline.

\subsubsection{Transformer models fail with limited data} The Sample Size task explores \new{methods}' performance on small datasets, a common occurrence in real world applications.
Generally, \hbert performs quite well. In this task, \hbert outperforms other \new{methods} when trained on the full 3,200 observation training set, but its bias and accuracy quickly deteriorate to worse than n-gram features when training data is reduced (Fig.~\ref{fig:results}j,k,l).
When data is especially scarce, practitioners should carefully consider the data-hungry nature of modern transformer architectures even when they are pretrained. A more sophisticated model is not always the best choice. Furthermore, transfer learning and other means to reduce training data requirements are an important area of future work for causal inference method developers.

\subsubsection{\new{Methods} estimate non-zero ATEs when the true ATE is zero} Alarmingly, in the Placebo Test, every \new{method} except \hbert's stratified estimate failed to include the (correct) null hypothesis ($ATE=0$) in their 95\% confidence intervals \new{across both datasets} (Fig.~\ref{fig:results}n,o), including high accuracy \new{methods} using bigram features (Fig.~\ref{fig:results}m).
This corresponds to incorrectly rejecting the null hypothesis of no causal treatment effect at the common p-value threshold of 0.05.
This result is of greatest concern, as 8/9 methods falsely claim a non-zero effect.

\subsubsection{\new{Text representations and propensity score models} have greater impact than estimators}
Each estimator evaluated produced overall similar results (Fig.~\ref{fig:matching_comparison}), with the quality of the propensity scores being far more impactful. Methods often cluster based on their text representations, with bigram representations generally performing better than unigrams, which generally perform better than LDA representations. There was generally little difference between Logistic Regression and Simple NN methods when trained on the same text representation.
However, the choice of an $ATE$ estimator is still important. IPTW is more sensitive to extreme or miscalibrated propensity scores. This is visible in the Strength of Selection Effect task, where the confidence intervals for IPTW are much larger than for the stratified estimator (Fig.~\ref{fig:results}h,i).

\subsection{Potential Threats to Evaluation Framework Validity}\label{sec:order_num_posts}
\new{
Many of our evaluation tasks (\sect\ref{sec:tasks}) append more synthetic posts to \texttt{class 1} histories than \texttt{class 2} histories. As a result, \texttt{class 1} histories will be 1-3 posts longer, in expectation, than those from \texttt{class 1}. Relative to the lengths of these histories (mean 41.1, standard deviation 19.3), this is a small difference, but in principle, a causal inference method may be able to pick up on the length of the history, an artifact of our framework, rather than the textual clues provided by the synthetic posts themselves. In practice, however, evidence from the Signal Intensity Task suggests that this is very unlikely, as every method tested, including \hbert, struggles with the to differentiate between histories by counting the number of synthetic posts (Fig.~\ref{fig:results}e,f). Thus, if we replaced a random post with a synthetic post, instead of appending, we would find very similar results. As methods improve and are better able to differentiate sequence lengths, such a random replacement strategy would be a reasonable and straightforward extension to our framework.
}

\new{
Additionally, our evaluation tasks (\sect\ref{sec:tasks}) always \textit{append} synthetic posts to users' histories. We conducted additional experiments using an `Order of Text' task (\appendixlink{Appendix}~\ref{appendix:order_of_text}) to evaluate widely used methods' ability to represent the order of posts in a user's history, and find that every method evaluated, including \hbert, completely fails to represent order. This is mostly a result of currently used text representations, primarily n-grams, which aggregate across histories by counting occurrences of tokens. For brevity, details of this task are included in \appendixlink{Appendix}~\ref{appendix:order_of_text}. We make this task public, along with all other tasks, to assist in the evaluation of future methods. We invite extensions and adaptions of our framework.
}

\section{Implications \& Conclusions}\label{sec:discussion}
Causal inferences are difficult to evaluate in the absence of ground truth causal effects -- a limitation of virtually all real world observational studies.
Despite this absence, we \textit{can} compare different methods' estimates and demonstrate that different methods regularly disagree with one another.

Empirical evaluation requires knowledge of the true treatment effects.
Our proposed evaluation framework is reflective of five key challenges for causal inference in natural language, \new{and is easily extensible to include different forms of confounders, different synthetic text content, and non-binary treatments and outcomes (\sect\ref{sec:generalizability}).}

\new{Our goal with this work is not to unilaterally pronounce one method as superior to another. Instead, we hope that methods developers and practitioners will comprehensively consider the challenges of making valid causal inferences we have described here, as well as assumptions they may be relying upon, and will use our framework to evaluate their methods empirically. To this end, }
we evaluate every commonly used propensity score method to produce key insights:

\noindent
For \textbf{methods developers}, we find that continued development of transformer-based models offers a promising path towards rectifying deficiencies of existing models. 
Models are needed that can effectively represent the order of text, variability in expression, and the counts of key tokens.
Given the limited availability of training data in many causal inference applications, more research is needed in adapting pretrained transformers to small data settings \cite{gururangan2020dont}.
We hope our public framework\footnote{\papersite} will provide a principled method for evaluating future NLP models for causal inference.

\noindent
For \textbf{practitioners}, we find that transformer-based methods such as \hbert, which we make publicly available,\footnotemark[9] perform the best in most cases except those with very limited data.
Propensity score models with high accuracy should be applied with great care, as this is likely indicative of a strong and unadjustable selection effect.
Many methods failed our placebo test by making false causal discoveries, a major problem \cite{aarts2015reproducability, freedman2015reproducability}.

\section*{Acknowledgments}

This research was supported in part by the Laboratory Directed Research and Development Program at Pacific Northwest National Laboratory, a multiprogram national laboratory operated by Battelle for the U.S. Department of Energy. 
This research was supported in part by Adobe Data Science Research Award, the Office for Naval Research (\#N00014-21-1-2154), NSF grant IIS-1901386, NSF grant CNS-2025022, the Bill \& Melinda Gates Foundation (INV-004841), a Microsoft AI for Accessibility grant, and a Garvey Institute Innovation grant.

{\small \bibliography{bibliography}}

\begin{thebibliography}{64}
\providecommand{\natexlab}[1]{#1}
\providecommand{\url}[1]{\texttt{#1}}
\providecommand{\urlprefix}{URL }
\expandafter\ifx\csname urlstyle\endcsname\relax
  \providecommand{\doi}[1]{doi:\discretionary{}{}{}#1}\else
  \providecommand{\doi}{doi:\discretionary{}{}{}\begingroup
  \urlstyle{rm}\Url}\fi

\bibitem[{Aarts et~al.(2015)Aarts, Anderson, Anderson, Attridge, Attwood, Axt,
  Babel, Bahnik, Baranski, Barnett-Cowan, Bartmess, Beer, Bell, Bentley, Beyan,
  Binion, Borsboom, Bosch, Bosco, and Penuliar}]{aarts2015reproducability}
Aarts, A.; Anderson, J.; Anderson, C.; Attridge, P.; Attwood, A.; Axt, J.;
  Babel, M.; Bahnik, S.; Baranski, E.; Barnett-Cowan, M.; Bartmess, E.; Beer,
  J.; Bell, R.; Bentley, H.; Beyan, L.; Binion, G.; Borsboom, D.; Bosch, A.;
  Bosco, F.; and Penuliar, M. 2015.
\newblock Estimating the Reproducibility of Psychological Science.
\newblock \emph{Science} 349.

\bibitem[{Abadie and Imbens(2016)}]{abadie2016matching}
Abadie, A.; and Imbens, G.~W. 2016.
\newblock Matching on the Estimated Propensity Score.
\newblock \emph{Econometrica} 84(2): 781--807.

\bibitem[{Arbour and Dimmery(2019)}]{arbour2019permutation}
Arbour, D.; and Dimmery, D. 2019.
\newblock Permutation Weighting.
\newblock \emph{arXiv:1901.01230} .

\bibitem[{Baumgartner et~al.(2020)Baumgartner, Zannettou, Keegan, Squire, and
  Blackburn}]{baumgartner2020pushshift}
Baumgartner, J.; Zannettou, S.; Keegan, B.; Squire, M.; and Blackburn, J. 2020.
\newblock The Pushshift Reddit Dataset.
\newblock \emph{arXiv:2001.08435} .

\bibitem[{Bhattacharya and Mehrotra(2016)}]{bhattacharya2016information}
Bhattacharya, P.; and Mehrotra, R. 2016.
\newblock The Information Network: Exploiting Causal Dependencies in Online
  Information Seeking.
\newblock CHIIR ’16.

\bibitem[{Blei, Ng, and Jordan(2003)}]{Blei2003LatentDA}
Blei, D.; Ng, A.; and Jordan, M.~I. 2003.
\newblock Latent Dirichlet Allocation.
\newblock \emph{J. Mach. Learn. Res.} 3: 993--1022.

\bibitem[{Brier(1950)}]{brier1950score}
Brier, G.~W. 1950.
\newblock Verification of Forecasts Expressed in Terms of Probability.
\newblock \emph{Monthly Weather Review} 78(1): 1--3.

\bibitem[{Chandrasekharan et~al.(2017)Chandrasekharan, Pavalanathan,
  Srinivasan, Glynn, Eisenstein, and Gilbert}]{chandrasekharan2017ban}
Chandrasekharan, E.; Pavalanathan, U.; Srinivasan, A.; Glynn, A.; Eisenstein,
  J.; and Gilbert, E. 2017.
\newblock You Can’t Stay Here: The Efficacy of Reddit’s 2015 Ban Examined
  Through Hate Speech.
\newblock \emph{CSCW} 1.

\bibitem[{Chandrasekharan et~al.(2018)Chandrasekharan, Samory, Jhaver, Charvat,
  Bruckman, Lampe, Eisenstein, and Gilbert}]{chandrasekharan2018internet}
Chandrasekharan, E.; Samory, M.; Jhaver, S.; Charvat, H.; Bruckman, A.; Lampe,
  C.; Eisenstein, J.; and Gilbert, E. 2018.
\newblock The Internet’s Hidden Rules: An Empirical Study of Reddit Norm
  Violations at Micro, Meso, and Macro Scales.
\newblock \emph{CSCW} 2.

\bibitem[{Chen, Montano-Campos, and Zadrozny(2020)}]{chen2020causal}
Chen, V.~Z.; Montano-Campos, F.; and Zadrozny, W. 2020.
\newblock Causal Knowledge Extraction from Scholarly Papers in Social Sciences.
\newblock \emph{arXiv:2006.08904} .

\bibitem[{Choudhury and De(2014)}]{choudhury2014self_disclosure}
Choudhury, M.~D.; and De, S. 2014.
\newblock Mental Health Discourse on reddit: Self-Disclosure, Social Support,
  and Anonymity.
\newblock In \emph{ICWSM}.

\bibitem[{Choudhury and Kiciman(2017)}]{choudhury2017language}
Choudhury, M.~D.; and Kiciman, E. 2017.
\newblock The Language of Social Support in Social Media and Its Effect on
  Suicidal Ideation Risk.
\newblock \emph{ICWSM} 2017: 32--41.

\bibitem[{Crump et~al.(2009)Crump, Hotz, Imbens, and Mitnik}]{crump2009dealing}
Crump, R.~K.; Hotz, V.~J.; Imbens, G.~W.; and Mitnik, O.~A. 2009.
\newblock {Dealing with limited overlap in estimation of average treatment
  effects}.
\newblock \emph{Biometrika} 96(1): 187--199.

\bibitem[{De~Choudhury et~al.(2016)De~Choudhury, Kiciman, Dredze, Coppersmith,
  and Kumar}]{dechoudhury2016mental_chi}
De~Choudhury, M.; Kiciman, E.; Dredze, M.; Coppersmith, G.; and Kumar, M. 2016.
\newblock Discovering Shifts to Suicidal Ideation from Mental Health Content in
  Social Media.
\newblock In \emph{CHI '16}.

\bibitem[{Devlin et~al.(2019)Devlin, Chang, Lee, and
  Toutanova}]{devlin-etal-2019-bert}
Devlin, J.; Chang, M.-W.; Lee, K.; and Toutanova, K. 2019.
\newblock {BERT}: Pre-training of Deep Bidirectional Transformers for Language
  Understanding.
\newblock In \emph{NAACL}.

\bibitem[{Dorie et~al.(2019)Dorie, Hill, Shalit, Scott, and
  Cervone}]{dorie2019acic}
Dorie, V.; Hill, J.; Shalit, U.; Scott, M.; and Cervone, D. 2019.
\newblock Automated versus Do-It-Yourself Methods for Causal Inference: Lessons
  Learned from a Data Analysis Competition.
\newblock \emph{Statist. Sci.} 34(1): 43--68.

\bibitem[{Eckles and Bakshy(2017)}]{eckles2017bias}
Eckles, D.; and Bakshy, E. 2017.
\newblock Bias and high-dimensional adjustment in observational studies of peer
  effects.
\newblock \emph{arXiv:1706.04692} .

\bibitem[{Egami et~al.(2018)Egami, Fong, Grimmer, Roberts, and
  Stewart}]{egami2018make}
Egami, N.; Fong, C.~J.; Grimmer, J.; Roberts, M.~E.; and Stewart, B.~M. 2018.
\newblock How to Make Causal Inferences Using Texts.
\newblock \emph{arXiv:1802.02163} .

\bibitem[{Falavarjani et~al.(2017)Falavarjani, Hosseini, Noorian, and
  Bagheri}]{falavarjani2017estimating}
Falavarjani, S. A.~M.; Hosseini, H.; Noorian, Z.; and Bagheri, E. 2017.
\newblock Estimating the Effect Of Exercising On Users’ Online Behavior.
\newblock In \emph{AAAI 2017}.

\bibitem[{Fong and Grimmer(2016)}]{fong2016discovery}
Fong, C.; and Grimmer, J. 2016.
\newblock Discovery of Treatments from Text Corpora.
\newblock In \emph{ACL '16}. Berlin, Germany: Association for Computational
  Linguistics.

\bibitem[{Freedman, Cockburn, and Simcoe(2015)}]{freedman2015reproducability}
Freedman, L.~P.; Cockburn, I.~M.; and Simcoe, T.~S. 2015.
\newblock The Economics of Reproducibility in Preclinical Research.
\newblock \emph{PLOS Biology} 13(6): 1--9.

\bibitem[{Gentzel, Garant, and Jensen(2019)}]{gentzel2019evaluating}
Gentzel, A.; Garant, D.; and Jensen, D. 2019.
\newblock The Case for Evaluating Causal Models Using Interventional Measures
  and Empirical Data.
\newblock In \emph{NeurIPS '19}.

\bibitem[{Gururangan et~al.(2020)Gururangan, Marasović, Swayamdipta, Lo,
  Beltagy, Downey, and Smith}]{gururangan2020dont}
Gururangan, S.; Marasović, A.; Swayamdipta, S.; Lo, K.; Beltagy, I.; Downey,
  D.; and Smith, N.~A. 2020.
\newblock Don't Stop Pretraining: Adapt Language Models to Domains and Tasks.

\bibitem[{Haimson(2018)}]{haimson2018transgender_disclosure}
Haimson, O.~L. 2018.
\newblock \emph{The Social Complexities of Transgender Identity Disclosure on
  Social Media.}
\newblock Ph.D. thesis, UC Irvine.

\bibitem[{H{\'a}jek(1970)}]{hajek1970}
H{\'a}jek, J. 1970.
\newblock A characterization of limiting distributions of regular estimates.
\newblock \emph{Zeitschrift f{\"u}r Wahrscheinlichkeitstheorie und Verwandte
  Gebiete} .

\bibitem[{Hill and Su(2013)}]{hill2013common_support}
Hill, J.; and Su, Y.-S. 2013.
\newblock Assessing lack of common support in causal inference using Bayesian
  nonparametrics: Implications for evaluating the effect of breastfeeding on
  children’s cognitive outcomes.
\newblock \emph{Ann. Appl. Stat.} 7(3).

\bibitem[{Hirano and Imbens(2004)}]{hirano2004continuous}
Hirano, K.; and Imbens, G.~W. 2004.
\newblock \emph{The Propensity Score with Continuous Treatments}, chapter~7,
  73--84.
\newblock John Wiley \& Sons, Ltd.

\bibitem[{Jensen(2019)}]{jensen2019comment}
Jensen, D. 2019.
\newblock Comment: Strengthening Empirical Evaluation of Causal Inference
  Methods.
\newblock \emph{Statist. Sci.} 34(1).

\bibitem[{Johansson, Shalit, and Sontag(2016)}]{johansson2016learning}
Johansson, F.; Shalit, U.; and Sontag, D. 2016.
\newblock Learning representations for counterfactual inference.
\newblock In \emph{ICML}.

\bibitem[{Joulin et~al.(2017)Joulin, Grave, Bojanowski, and
  Mikolov}]{joulin-etal-2017-bag}
Joulin, A.; Grave, E.; Bojanowski, P.; and Mikolov, T. 2017.
\newblock Bag of Tricks for Efficient Text Classification.
\newblock In \emph{Proceedings of the 15th Conference of the {E}uropean Chapter
  of the Association for Computational Linguistics: Volume 2, Short Papers},
  427--431. Valencia, Spain: Association for Computational Linguistics.
\newblock \urlprefix\url{https://www.aclweb.org/anthology/E17-2068}.

\bibitem[{Kang and Schafer(2007)}]{kang2007demystifying}
Kang, J. D.~Y.; and Schafer, J.~L. 2007.
\newblock Demystifying Double Robustness: A Comparison of Alternative
  Strategies for Estimating a Population Mean from Incomplete Data.
\newblock \emph{Statist. Sci.} 22(4): 523--539.

\bibitem[{Keith, Jensen, and O'Connor(2020)}]{keith2020review}
Keith, K.~A.; Jensen, D.; and O'Connor, B. 2020.
\newblock Text and Causal Inference: A Review of Using Text to Remove
  Confounding from Causal Estimates.
\newblock In \emph{ACL}.

\bibitem[{Kiciman, Counts, and Gasser(2018)}]{kiciman2018UsingLS}
Kiciman, E.; Counts, S.; and Gasser, M. 2018.
\newblock Using Longitudinal Social Media Analysis to Understand the Effects of
  Early College Alcohol Use.
\newblock In \emph{ICWSM}.

\bibitem[{Kingma and Ba(2014)}]{kingma2014adam}
Kingma, D.~P.; and Ba, J. 2014.
\newblock Adam: A Method for Stochastic Optimization.
\newblock \emph{arXiv:1412.6980} .

\bibitem[{Landeiro and Culotta(2016)}]{landeiro2016robust}
Landeiro, V.; and Culotta, A. 2016.
\newblock Robust Text Classification in the Presence of Confounding Bias.
\newblock \emph{AAAI} .

\bibitem[{Lee, Lessler, and Stuart(2011)}]{lee2011weight}
Lee, B.~K.; Lessler, J.; and Stuart, E.~A. 2011.
\newblock Weight Trimming and Propensity Score Weighting.
\newblock \emph{PLOS ONE} 6(3).

\bibitem[{Li, Thomas, and Li(2018)}]{li2018addressing}
Li, F.; Thomas, L.~E.; and Li, F. 2018.
\newblock {Addressing Extreme Propensity Scores via the Overlap Weights}.
\newblock \emph{Am. J. Epidemiol} 188(1): 250--257.

\bibitem[{Lin et~al.(2019)Lin, Merchant, Sarkar, and
  D'Amour}]{lin2019universal}
Lin, A.; Merchant, A.; Sarkar, S.~K.; and D'Amour, A. 2019.
\newblock Universal Causal Evaluation Engine: An API for empirically evaluating
  causal inference models.
\newblock In Le, T.~D.; Li, J.; Zhang, K.; Cui, E. K.~P.; and Hyvärinen, A.,
  eds., \emph{Proceedings of Machine Learning Research}, volume 104 of
  \emph{Proceedings of Machine Learning Research}, 50--58. Anchorage, Alaska,
  USA: PMLR.
\newblock \urlprefix\url{http://proceedings.mlr.press/v104/lin19a.html}.

\bibitem[{Mani and Cooper(2000)}]{mani2000discovery}
Mani, S.; and Cooper, G.~F. 2000.
\newblock Causal discovery from medical textual data.
\newblock \emph{Proceedings. AMIA Symposium} .

\bibitem[{Mayer et~al.(2019)Mayer, Wager, Gauss, Moyer, and
  Josse}]{mayer2019doubly}
Mayer, I.; Wager, S.; Gauss, T.; Moyer, J.-D.; and Josse, J. 2019.
\newblock Doubly robust treatment effect estimation with missing attributes.
\newblock \emph{arXiv:1910.10624} .

\bibitem[{Medvedev, Lambiotte, and Delvenne(2019)}]{medvedev2019anatomy}
Medvedev, A.~N.; Lambiotte, R.; and Delvenne, J.-C. 2019.
\newblock The Anatomy of Reddit: An Overview of Academic Research.
\newblock \emph{Springer Proceedings in Complexity} 183–204.

\bibitem[{Mirza and Tonelli(2016)}]{mirza2016catena}
Mirza, P.; and Tonelli, S. 2016.
\newblock {CATENA}: {CA}usal and {TE}mporal relation extraction from {NA}tural
  language texts.
\newblock In \emph{COLING '16}. Osaka, Japan.

\bibitem[{Mozer et~al.(2020)Mozer, Miratrix, Kaufman, and
  Jason~Anastasopoulos}]{mozer2020matching}
Mozer, R.; Miratrix, L.; Kaufman, A.~R.; and Jason~Anastasopoulos, L. 2020.
\newblock Matching with Text Data: An Experimental Evaluation of Methods for
  Matching Documents and of Measuring Match Quality.
\newblock \emph{Political Analysis} 1–24.

\bibitem[{Olteanu, Varol, and Kiciman(2017)}]{olteanu2017distilling}
Olteanu, A.; Varol, O.; and Kiciman, E. 2017.
\newblock Distilling the Outcomes of Personal Experiences: A Propensity-Scored
  Analysis of Social Media.
\newblock In \emph{CSCW}.

\bibitem[{Pearl(1995)}]{pearl1995causal}
Pearl, J. 1995.
\newblock Causal Diagrams for Empirical Research.
\newblock \emph{Biometrika} 82(4).
\newblock ISSN 00063444.

\bibitem[{Pham and Shen(2017)}]{pham2017deep}
Pham, T.~T.; and Shen, Y. 2017.
\newblock A Deep Causal Inference Approach to Measuring the Effects of Forming
  Group Loans in Online Non-profit Microfinance Platform.

\bibitem[{Pryzant et~al.(2018)Pryzant, Shen, Jurafsky, and
  Wagner}]{pryzant2018deconfounded}
Pryzant, R.; Shen, K.; Jurafsky, D.; and Wagner, S. 2018.
\newblock Deconfounded Lexicon Induction for Interpretable Social Science.
\newblock In \emph{NAACL}.

\bibitem[{Roberts, Stewart, and Nielsen(2020)}]{roberts2018adjusting}
Roberts, M.~E.; Stewart, B.~M.; and Nielsen, R.~A. 2020.
\newblock Adjusting for confounding with text matching.
\newblock \emph{American Journal of Political Science} .

\bibitem[{Rosenbaum(2010)}]{rosenbaum_2010}
Rosenbaum, P.~R. 2010.
\newblock \emph{Design of Observational Studies}.
\newblock Springer.

\bibitem[{Rubin(2005)}]{rubin_2005_potential_outcomes}
Rubin, D.~B. 2005.
\newblock Causal Inference Using Potential Outcomes.
\newblock \emph{Journal of the American Statistical Association} 100(469):
  322--331.
\newblock \doi{10.1198/016214504000001880}.
\newblock \urlprefix\url{https://doi.org/10.1198/016214504000001880}.

\bibitem[{Saha et~al.(2019)Saha, Sugar, Torous, Abrahao, Kiciman, and
  Choudhury}]{saha2019ASM}
Saha, K.; Sugar, B.; Torous, J.~B.; Abrahao, B.~D.; Kiciman, E.; and Choudhury,
  M.~D. 2019.
\newblock A Social Media Study on the Effects of Psychiatric Medication Use.
\newblock \emph{ICWSM} 13.

\bibitem[{Schuler and Rose(2017)}]{schuler2017tmle}
Schuler, M.; and Rose, S. 2017.
\newblock Targeted Maximum Likelihood Estimation for Causal Inference in
  Observational Studies.
\newblock \emph{American Journal of Epidemiology} 185: 65–73.

\bibitem[{Sridhar et~al.(2018)Sridhar, Springer, Hollis, Whittaker, and
  Getoor}]{sridhar2018estimating}
Sridhar, D.; Springer, A.; Hollis, V.; Whittaker, S.; and Getoor, L. 2018.
\newblock Estimating Causal Effects of Exercise from Mood Logging Data.
\newblock \emph{FAIM'18 Workshop on CausalML} .

\bibitem[{Talwar et~al.(2019)Talwar, Dhir, Kaur, Zafar, and
  Alrasheedy}]{talwar2019fake_news}
Talwar, S.; Dhir, A.; Kaur, P.; Zafar, N.; and Alrasheedy, M. 2019.
\newblock Why do people share fake news? Associations between the dark side of
  social media use and fake news sharing behavior.
\newblock \emph{Journal of Retailing and Consumer Services} 51: 72--82.

\bibitem[{Tan, Lee, and Pang(2014)}]{tan2014effect}
Tan, C.; Lee, L.; and Pang, B. 2014.
\newblock The effect of wording on message propagation: Topic- and
  author-controlled natural experiments on Twitter.
\newblock In \emph{ACL}.

\bibitem[{Veitch, Sridhar, and Blei(2019)}]{veitch2019embeddings}
Veitch, V.; Sridhar, D.; and Blei, D.~M. 2019.
\newblock Using Text Embeddings for Causal Inference.
\newblock \emph{arXiv:1905.12741} .

\bibitem[{Wang et~al.(2013)Wang, Cai, Li, Jiang, Wang, Song, and
  Xia}]{wang2013caliper}
Wang, Y.; Cai, H.; Li, C.; Jiang, Z.; Wang, L.; Song, J.; and Xia, J. 2013.
\newblock Optimal Caliper Width for Propensity Score Matching of Three
  Treatment Groups: A Monte Carlo Study.
\newblock \emph{PLOS ONE} 8: 1--7.

\bibitem[{Wang and Culotta(2019)}]{wang2019lexical_choice}
Wang, Z.; and Culotta, A. 2019.
\newblock When Do Words Matter? Understanding the Impact of Lexical Choice on
  Audience Perception Using Individual Treatment Effect Estimation.
\newblock In \emph{AAAI}.

\bibitem[{Wood-Doughty, Shpitser, and
  Dredze(2018)}]{wood_doughty2018challenges}
Wood-Doughty, Z.; Shpitser, I.; and Dredze, M. 2018.
\newblock Challenges of Using Text Classifiers for Causal Inference.
\newblock In \emph{EMNLP}.

\bibitem[{Yang and Ding(2018)}]{yang2018asymptotic}
Yang, S.; and Ding, P. 2018.
\newblock {Asymptotic inference of causal effects with observational studies
  trimmed by the estimated propensity scores}.
\newblock \emph{Biometrika} .

\bibitem[{Yu and Mu{\~n}oz-Justicia(2020)}]{Yu2020ABO}
Yu, J.; and Mu{\~n}oz-Justicia, J. 2020.
\newblock A Bibliometric Overview of Twitter-Related Studies Indexed in Web of
  Science.
\newblock \emph{Future Internet} 12: 91.

\bibitem[{Zhang et~al.(2018)Zhang, Chang, Danescu-Niculescu-Mizil, Dixon, Hua,
  Taraborelli, and Thain}]{zhang2018conversations}
Zhang, J.; Chang, J.; Danescu-Niculescu-Mizil, C.; Dixon, L.; Hua, Y.;
  Taraborelli, D.; and Thain, N. 2018.
\newblock Conversations Gone Awry: Detecting Early Signs of Conversational
  Failure.
\newblock In \emph{ACL}.

\bibitem[{Zhang, Mullainathan, and
  Danescu-Niculescu-Mizil(2020)}]{zhang2020quantifying}
Zhang, J.; Mullainathan, S.; and Danescu-Niculescu-Mizil, C. 2020.
\newblock Quantifying the Causal Effects of Conversational Tendencies.
\newblock \emph{Proc. ACM Hum.-Comput. Interact.} 4(CSCW2).
\newblock \doi{10.1145/3415202}.
\newblock \urlprefix\url{https://doi.org/10.1145/3415202}.

\bibitem[{Zhang, Wei, and Zhou(2019)}]{zhang-etal-2019-hibert}
Zhang, X.; Wei, F.; and Zhou, M. 2019.
\newblock {HIBERT}: Document Level Pre-training of Hierarchical Bidirectional
  Transformers for Document Summarization.
\newblock In \emph{ACL}.

\end{thebibliography}


\appendix
\clearpage
\onecolumn

\begin{center}
\LARGE
{\bf
Adjusting for Confounders with Text: Challenges and an Empirical Evaluation Framework for Causal Inference 
} \\
Appendices
\end{center}

\section{Theoretical Bounds}
\label{appendix:bounds}\label{sec:bounds}

We leverage recent results of \citet{arbour2019permutation} to bound the expected bias of the ATE, $\bar{Y}[T=1]-\bar{Y}[T=0]$, by considering the weighted risk of the propensity score:

\vspace{0.5\baselineskip}
 \noindent 
    { \small $
    \left| \mathbb{E}\left[\hat{Y}(T)\right] - \mathbb{E}\left[ \vphantom{\hat{Y}(T)} Y(T)\right] \right| 
    \leq \left|  \mathbb{E}\left[\frac{Y}{p(T|X)}\frac{S(\hat{p}(T|X), p(T|X))}{\hat{p}(T|X)^2} \right]\right|
$
} 

\vspace{0.5\baselineskip}
\noindent
where $\hat{p}$ and $p$ are the estimated and true propensity score, and $S$ is the Brier score~\shortcite{brier1950score}.
Conceptually, this bound suggests that the bias grows as a function of the Brier score between estimated and true propensity score (numerator), and the inverse of the squared estimate of the propensity score, significantly penalizing very small scores.


\subsubsection{Findings}
We compute these bounds using the estimated propensity score and find that they are largely uninformative in practice.
In 250/252 cases, the empirical confidence interval (Fig.~\ref{fig:results}) provides a tighter bound than the theoretical bound, and in 230/252 cases the Unadjusted Estimator also provides a tighter bound than the theoretical bound.
These results again highlight the importance of the principled empirical evaluation framework presented here.

\subsubsection{Details of Derivation}
The central challenge is estimating the error of the counterfactual quantities, $Y(1)$, and $Y(0)$. 
Recall that in the case of weighting estimators, when the true propensity score~($p(\cdot)$) is available, these are estimated as
$\mathbb{E}\left[ y(T)\right] = \mathbb{E}\left[\frac{Y}{p(T)}\right]$, where $y$ is the observed outcome. 
For the problem addressed in this paper, the propensity must be estimated. 
Estimating the error for each potential outcome under an estimated propensity score results in a bias of  
 
\vspace{0.5\baselineskip}
 \noindent
\begin{centering}
    { \small $
    \left| \mathbb{E}\left[\hat{Y}(T)\right] - \mathbb{E}\left[ \vphantom{\hat{Y}(T)} Y(T)\right] \right| = \left|  \mathbb{E}\left[\frac{Y}{p(T|X)}\right]  -  \mathbb{E}\left[\frac{Y}{\hat{p}(T|X)}\right]\right|
$
}
\end{centering}
\vspace{0.5\baselineskip}
 
\noindent
following Proposition 1 of \citet{arbour2019permutation}.

More concretely, an empirical upper bound can be obtained for Equation \ref{eq:propbias} given a lower bound on the true propensity score. 
Specifically, replacing the $p$ with the lower bound and using the weighted cross-validated Brier score will provide a conservative bound on the bias of the counterfactual.
This bound can be tightened with further assumptions, for example by assuming instance level bounds on $p$ instead of a global bound. 
Balancing weights may also be used to estimate the bias directly using only empirical quantities~\citep{arbour2019permutation}. 

Note that due to the evaluation framework in this paper, the true propensity score $p$ is known, and therefore we do not need to apply loose bounds.

\vspace{-\baselineskip}
\begin{align}
    \left| \mathbb{E}\left[\hat{Y}(T)\right] - \mathbb{E}\left[ \vphantom{\hat{Y}(T)} Y(T)\right] \right|
   &  = \left|  \mathbb{E}\left[\frac{y}{p(T)}  -  \frac{y}{p(T) + (\hat{p}(T) - p(T))}\right]\right| \nonumber \\
  &   = \left|  \mathbb{E}\left[\frac{y}{p(T)}  -  \frac{y}{p(T) + (\hat{p}(T) - p(T))}\right]\right| \nonumber  \\
  &  = \left|  \mathbb{E}\left[\frac{y(1 + \frac{1}{p}(\hat{p}(T) - p(T)))}{p(T)(1 + \frac{1}{p}(\hat{p}(T) - p(T)))}  -  \frac{y}{p(T) + (\hat{p}(T) - p(T))}\right]\right| \nonumber \\
  &  = \left|  \mathbb{E}\left[\frac{y + \frac{y}{p}(\hat{p}(T) - p(T))}{p(T) +  (\hat{p}(T) - p(T))}  -  \frac{y}{p(T) + (\hat{p}(T) - p(T))}\right]\right|  \nonumber \\
  &  = \left|  \mathbb{E}\left[\frac{y}{p}\frac{(\hat{p}(T) - p(T))}{\hat{p}(T)} \right]\right| \nonumber \\
  &  \leq   \left|  \mathbb{E}\left[\frac{y}{p}\frac{(\hat{p}(T) - p(T))^2}{\hat{p}(T)^2} \right]\right| \nonumber \\
  &  \leq \left|  \mathbb{E}\left[\frac{y}{p(T)}\frac{S(\hat{p}(T), p(T))}{\hat{p}(T)^2} \right]\right|
    \label{eq:propbias}
\end{align}
\vspace{\baselineskip}

\newpage

After obtaining the bounds on the individual counterfactual quantities, the corresponding lower and upper bias bounds on the average treatment effect can be constructed by considering

\begin{align}
    \label{eq:b1}
    &\hat{Y}(0) + \left|  \mathbb{E}\left[\frac{y}{p(T=0|X)}\frac{S(\hat{p}(0|X), p(0|X))}{\hat{p}(T=0|X)^2} \right]\right| \\
    &\hat{Y}(1) - \left|  \mathbb{E}\left[\frac{y}{p(T=1 \mid X)}\frac{S(\hat{p}(T = 1 \mid X), p(T= 1 \mid X))}{\hat{p}(T = 1 \mid X)^2} \right]\right|
\end{align}

\noindent 
and

\begin{align}
\label{eq:b2}
    &\hat{Y}(0) - \left|  \mathbb{E}\left[\frac{y}{p(T=0|X)}\frac{S(\hat{p}(T=0|X), p(0))}{\hat{p}(T=0|X)^2} \right]\right| \\
    &\hat{Y}(1) + \left|  \mathbb{E}\left[\frac{y}{p(T=1|X)}\frac{S(\hat{p}(T=1|X), p(T=1|X))}{\hat{p}(T=1|X)^2} \right]\right|
\end{align}

\noindent 
respectively.

\clearpage
\section{Moderation and Gender Experiments -- Data Collection Details}
\label{appendix:motivation_experiments}
\subsection{Moderation Experiment}
In the Moderation Experiment, we test if having a post removed by a moderator impacts the amount a user later posts to the same community. For this experiment, we use 13,786 public Reddit histories (all of which contain more than 500 tokens) from users in \texttt{/r/science} from 2015-2017 who had a not had a post removed prior to 2018.
Our treated users are those who \textit{have} had a post removed in 2018.
Out untreated users are those who have \textit{not} had a post removed in 2018 (nor before).
The outcome of interest is the number of posts they made in 2019.

To determine which users have had posts removed, we utilize the Pushshift Reddit API \cite{baumgartner2020pushshift}.
The data acessible via this API, in combination with publicly available Pushshift dump archives, allow us to compare two snapshots of each Reddit post: one snapshot made within a few seconds of posting, and one made approximately 2 months later.
By comparing these two versions, we can tell a) which user made the post, and b) if it was removed.
This approach is similar to that of \citet{chandrasekharan2018internet}.

This experiment mimics the setup in \citet{dechoudhury2016mental_chi}, where each user is represented by their entire Reddit comment history within specific subreddits.
While \cite{dechoudhury2016mental_chi} has been influential in our work, their dataset is not public, and publicly available comparable data contains only a relatively small set of Reddit users, leading to underpowered experiments with large, uninformative confidence intervals that fail to reproduce the findings in the original paper.

\subsection{Gender Experiment}
In the Gender Experiment, we use the dataset made public by \citet{veitch2019embeddings}, which consists of single posts from three subreddits: \texttt{/r/okcupid}, \texttt{/r/childfree}, and \texttt{/r/keto}. 
Each post is annotated with the gender (male or female) of the poster, which is considered the treatment.
The outcome is the score of the post (number of `upvotes' minus number of `downvotes').

\clearpage
\section{Templates for Synthetic Posts}
\label{appendix:templates}
As described in the `Synthetic Posts' section, synthetic \textit{sickness}, \textit{social isolation}, and \textit{death} posts are used to generate our evaluation tasks.
These synthetic posts are selected and inserted into social media histories of real world users by randomly sampling a template and word pair, or, in the case of Social Isolation Posts, by randomly sampling a complete post.

\subsection{Sickness Posts}
Sickness Posts are created by randomly sampling a Sickness Word and inserting it into a randomly sampled Sickness Template.
 
\noindent
Sickness Templates are sampled from:
{\small
    \begin{flalign*}
    \{
    &\texttt{The doctor told me I have \textbf{x}}, &\\
    &\texttt{I was at the hospital earlier and I have \textbf{x}.}, &\\
    &\texttt{I got diagnosed with \textbf{x} last week.}, &\\
    &\texttt{Have anyone here dealt with \textbf{x}? I just got diagnosed.}, &\\
    &\texttt{How should I handle a \textbf{x} diagnosis?}, &\\
    &\texttt{How do I tell my parents I have \textbf{x}?} &
    \}
    \end{flalign*}
}%
Sickness Words are sampled from
{\small
\{\texttt{cancer},
\texttt{ leukemia},
\texttt{ HIV},
\texttt{ AIDS},
\texttt{ Diabetes},
\texttt{ lung cancer},
\texttt{ stomach cancer},
\texttt{ skin cancer},
\texttt{ parkinson`s}\}
}

\subsection{Social Isolation Posts}
Social Isolation Posts are randomly sampled from the following set of complete synthetic posts:
{\small
    \begin{flalign*}
    \{
    &\texttt{My friends stopped talking to me.}, &\\
    &\texttt{My wife just left me.}, &\\
    &\texttt{My parents kicked me out of the house today.}, &\\
    &\texttt{I feel so alone, my last friend said they needed to stop seeing me.}, &\\
    &\texttt{My partner decided that we shouldn`t talk anymore last night.}, &\\
    &\texttt{My folks just cut me off, they won`t talk to me anymore.}, &\\
    &\texttt{I just got a message from my brother that said he can`t talk to me anymore. He was my} &\\
    &        \hspace{30pt}\texttt{last contact in my family.}, &\\
    &\texttt{My last friend at work quit, now there`s no one I talk to regularly.}, &\\
    &\texttt{I tried calling my Mom but she didn't pick up the phone. I think my parents may be done} &\\
    &        \hspace{30pt}\texttt{with me.}, &\\
    &\texttt{I got home today and my partner was packing up to leave. Our apartment feels so empty} &\\
    &        \hspace{30pt}\texttt{now.} &
    \}
    \end{flalign*}
}

\subsection{Death Posts}
Death Posts are created by randomly sampling a Death Word and inserting it into a Death Template.

\noindent
Death Templates are sampled from:
{\small
    \begin{flalign*}
    \{
    &\texttt{My \textbf{x} just died}, &\\
    &\texttt{I just found out my \textbf{x} died}, &\\
    &\texttt{My \textbf{x} died last weekend}, &\\
    &\texttt{What do you do when your \textbf{x} dies? This happened to me.}, &\\
    &\texttt{Has anyone else had a \textbf{x} die recently?}, &\\
    &\texttt{I lost my \textbf{x} yesterday.}, &\\
    &\texttt{My \textbf{x} passed away recently.}, &\\
    &\texttt{I am in shock. My \textbf{x} is gone.} &
    \}
    \end{flalign*}
}%
Death Words are sampled from
{\small
\{\texttt{Mom},
\texttt{ Mother},
\texttt{ Mama},
\texttt{ Father},
\texttt{ Dad},
\texttt{ Papa},
\texttt{ Brother},
\texttt{ Wife},
\texttt{ girlfriend},
\texttt{ partner},
\texttt{ spouse},
\texttt{ husband},
\texttt{ son},
\texttt{ daughter},
\texttt{ best friend}\}
}

\clearpage
\new{
\section{Vocabulary Sizes for N-Gram Methods}\label{appendix:vocab_size}
We report the sizes of the vocabularies used by the uni- and bi-gram text representations across the five tasks in the proposed evaluation framework. The vocabulary size varies slightly for each task due to different synthetic posts used; the maximum is reported here.
\begin{table}[h]
    \centering
    \begin{tabular}{l|l|r}
    \textbf{Real Data Source} & \textbf{N-Gram type} & \textbf{Maximum Vocabulary Size} \\
    \hline
    Reddit  & unigram &  24,164  \\
    Reddit  & bigram  & 125,319  \\
    Twitter & unigram &  23,791  \\
    Twitter & bigram  &  95,934  \\
    \end{tabular}
    \caption{Vocabulary sizes for each n-gram type and real world data source.}
    \label{tab:voca_size}
\end{table}
}

\clearpage
\section{Model Implementation, Tuning, and Parameters}\label{appendix:model_implementation}
\subsection{\hbert Architecture}\label{appendix:sherbert_architecture}
\begin{figure}[!ht]
    \centering
    \includegraphics[trim=1.5cm 0 0 0, clip, width=1.0\columnwidth]{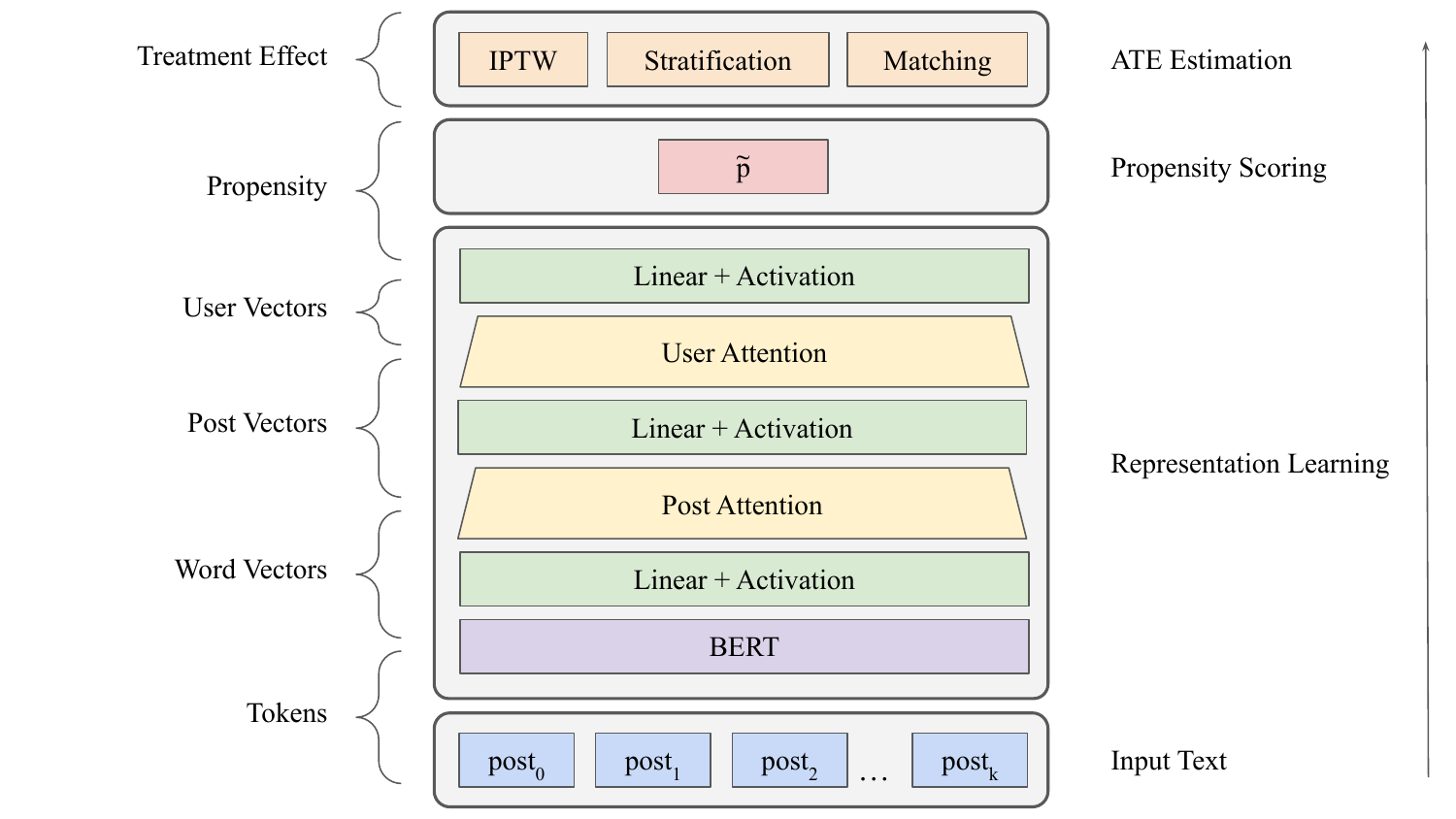}
    \caption{The complete ATE estimation pipeline, with tokens input at the bottom, and an estimate propensity at the top. ATE estimates are computed with IPTW, Stratification, and Matching based on models' propensity scores. This example is instantiated using \hbert and detailing its hierarchical architecture. In this pipeline, other propensity score models could replace the `Representation Learning' box (\eg, Bag-of-n-grams with Logistic Regression).}
    \label{fig:herbert}
\end{figure}

Fig.~\ref{fig:herbert} depicts the architecture of \hbert as part of the broader ATE estimation pipeline. 

Our work attempts to expand the success of large pretrained transformers to long history length using a hierarchical attention, which is a problem also explored by the HIBERT model in \citet{zhang-etal-2019-hibert}.
Essentially, \hbert differs from HIBERT in that \hbert trains a light-weight hierarchical attention on top of the pretrained BERT model \cite{devlin-etal-2019-bert} whereas HIBERT is trained from scratch.
This results in a relatively simple training procedure for \hbert, and lighter limitations on history length, both at the local (50 words for HIBERT v. 512 wordpiece tokens for \hbert) and global (30 sentences for HIBERT v. 60 for \hbert) scales.
This reflects differing tradeoffs; where HIBERT has a more sophisticated attention mechanism for combining local and global information, \hbert sacrifices some complexity for fast and simple training and longer text histories. 

\subsection{Practicality of Models}\label{appendix:practicality}
\hbert trades-off practicality for performance in comparison to simpler models. For instance, in most experiments we found \hbert takes 10 - 12 hours to train, sometimes requiring multiple starts to converge to a reasonable model. In contrast, training all other models collectively requires less than 1 hour. Further, the performance of \hbert sharply suffered as the number of users was reduced (Fig.~\ref{fig:results}j). While effectively training \hbert on 1 GPU (Tesla V100) in under 24 hours is quite practical compared to contemporary text pretraining regimes \cite{devlin-etal-2019-bert}, these issues should be considered when deciding on a causal text model.

\subsection{Hyperparameters}
\label{appendix:hyperparams}
A complete description of parameters and hyperparameters are available at the paper website.\footnote{\papersite} Basic details are included here.

In producing n-gram features, a count threshold of 10 is used to filter out low frequency words, and word tokenization is done using the NLTK word tokenizer. In producing LDA features, we use the Scikit Learn implementation, with 20 topics. To produce BERT word embedding features, we use the uncased model of the `base' size. 

All models use the Adam optimizer \cite{kingma2014adam}, with various learning rates decided empirically depending on model and task to maximize treatment accuracy on the validation set.

For the simple neural network model, we use a hidden size of 10. For \hbert, we use hidden sizes of 1000 and dot-product attention.

\clearpage
\section{Additional Estimators and Metrics}
In order to further detail our findings, we include several additional ATE estimators and metrics for evaluation (\sect\ref{sec:metrics}).

\subsection{Matching Estimator}\label{appendix:matching}
Matching can be considered as a special case of stratification, where each strata contains only one treated user.
As our treated and untreated groups are approximately balanced, we implement 1:1 matching, where each treated user is matched to exactly one untreated user.

While there are many implementations of matching, we implement matching \textit{with} replacement, as in \citet[pg.~784]{abadie2016matching}:
\begin{align*}
   \widehat{ATE}_\texttt{match} =  \frac{1}{n} \sum_{i=1}^n \left( 2 T_i - 1 \right) \left( Y_i - Y_j \right)
\end{align*}
\noindent where $j$ is the matched observation, \ie $j = \min_{j \in \{1\ldots n\}} |\hat{p}(\mathbf{X}_i) - \hat{p}(\mathbf{X}_i) |$ where $T_i \neq T_j$.

\noindent
A recent evaluation of matching techniques for text found no significant difference in match quality between matches produced with and without replacement \cite{mozer2020matching}.
We use a caliper value of $.2 \times$ the standard deviation of propensity scores in the population, as was found to perform the best by \citet{wang2013caliper} and recommended by \citet[pg.~251]{rosenbaum_2010}.

\begin{figure*}
    \centering
    \vspace{-6pt}
    \centerline{\includegraphics[height=.97\textheight]{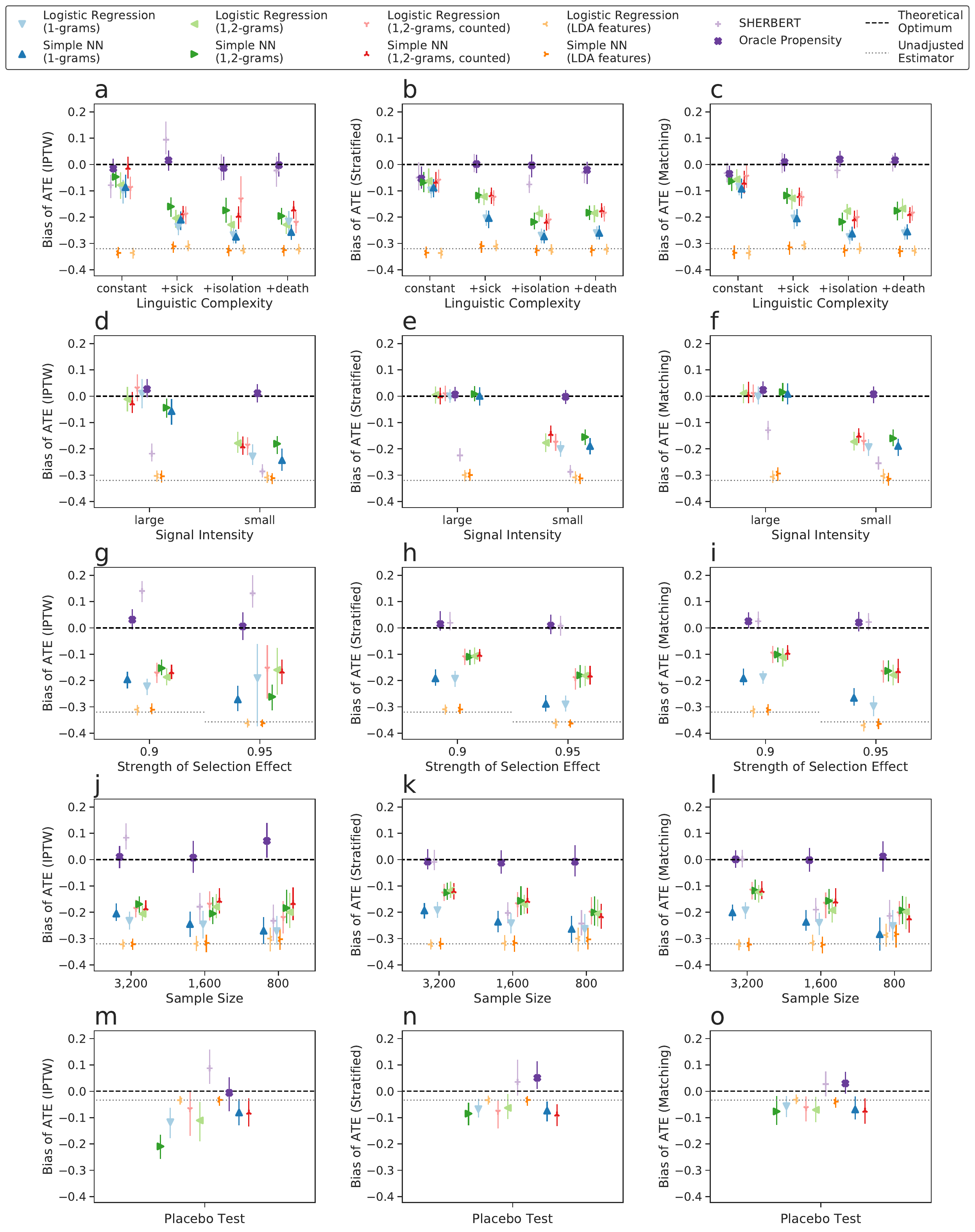}}
    \caption{Comparison of bias computed using IPTW, Stratification, and Propensity Score Matching, for each task, \new{computed using Reddit data}. Note that matching produces extremely similar results to stratification.} 
    \label{fig:matching_comparison}
\end{figure*}

For each of the five tasks, the matching estimator produces results extremely similar to those of the stratified estimator (Fig.~\ref{fig:matching_comparison}).

\subsection{Mean Squared Error of IPTW and Spearman Correlation}\label{appendix:additional_metrics}
In addition to Treatment Accuracy and Bias, we computed the Mean Squared Error (MSE) of the Inverse Probability of Treatment Weights, and the Spearman Correlation of propensity scores.

\textbf{Mean Squared Error of IPTW} shows the absolute error in the calibration of a models' causal weights:
\begin{align*}
&MSE_{IPTW} = 
        &\sum_{i=1}^n \left(\frac{\left[\sum_{j=1}^n
        \frac{1}{\hat{p}_{T_j} (\mathbf{X}_j)}\right]^{-1}}{\hat{p}_{T_i} (\mathbf{X}_i) }
        -  \frac{\left[\sum_{j=1}^n \frac{1}{p_{T_j} (\mathbf{X}_j)}\right]^{-1}}{p_{T_i} (\mathbf{X}_i) } \right ) ^2
\end{align*}
Notation is the same as in the `Average Treatment Effect Estimators' section of the main paper, with the addition of $p$ as true propensity, which is known in our semi-synthetic tasks.
The MSE is fairly correlated with the Treatment Accuracy, with MSE increasing as accuracy decreases as the tasks become more difficult.
This is especially evident in Fig.~\ref{appendix:additional_metrics}b,k.

\begin{figure*}
    \centering
    \vspace{-6pt}
    \centerline{\includegraphics[height=.97\textheight]{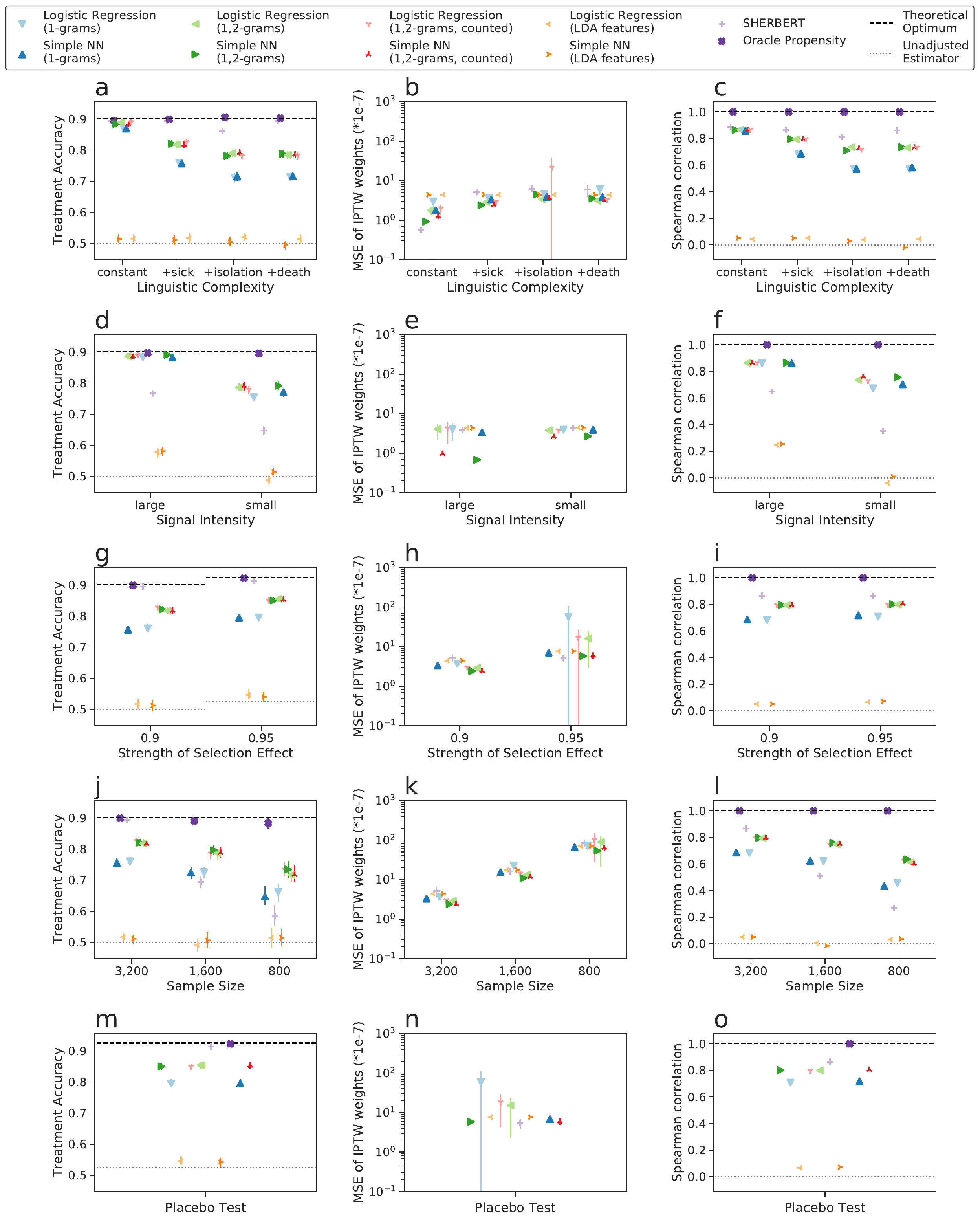}}
    \caption{Treatment Accuracy, Mean Squared Error, and Spearman Correlation for each task, \new{computed using Reddit data}. Spearman Correlation varies directly with Treatment Accuracy, whereas Mean Squared Error increases as accuracy falls.}
    \label{fig:additional_metrics}
\end{figure*}

\textbf{Spearman Correlation} instead shows the relative calibration of a models' propensity scores.
Propensity scores may have poor absolute calibration, but still have meaningful relative ordering, in which case the Spearman Rank Correlation is close to its maximum value of 1.
The Spearman Correlation coefficient is simply the Pearson correlation coefficient between the rank variables for the estimated and actual propensity scores.
We find that Spearman Correlation is also quite correlated with the Treatment Accuracy (Fig.~\ref{fig:additional_metrics}c,i), \new{and as such, is no more useful than Treatment Accuracy at predicting bias.}

\newpage

\newpage

\clearpage
\section{Twitter Results}\label{appendix:twitter_results}
\new{
We evaluate each text representation and propensity score model on two versions of each of the five tasks, one version generated using Reddit data, and one version generated using Twitter data (for details on data collection, see \sect\ref{sec:real_data}).
Language used on Twitter differs from language used on Reddit, and as a result, the difficulty of our tasks varies slightly from platform to platform. However, the semi-synthetic nature of our tasks leads us to expect the differences in downstream evaluation metrics to be negligible, a hypothesis supported by our findings.
Overall, the results are extremely similar across the two datasets, with Treatment Accuracies within 1.9 percentage points of one another. Therefore, for brevity, we focus on the Reddit results in the main body of the paper. Here, we report Twitter results and compare them to the Reddit results.
}

\begin{table}[h]
    \centering
    
\begin{tabular}{l|rrr}
{} &  \textbf{Mean Absolute Difference} &  \textbf{Mean Twitter:Reddit Ratio} &  \textbf{Pearson's r} \\
\hline
\textbf{Test Accuracy   } &                     0.019 &                      1.006 &        0.957 \\
\textbf{ATE (Stratified)} &                     0.032 &                      1.042 &        0.946 \\
\textbf{ATE (IPTW)      } &                     0.052 &                      1.157 &        0.887 \\
\end{tabular}

\caption{Comparison of Reddit and Twitter results. Note that results are highly similar across the two datasets with mean absolute differences of 0.019 for Treatment Accuracy for values between 0 and 1, mean absolute differences of 0.032 for the $ATE$ estimate using stratification,and  0.052 for the $AT$E estimate using IPTW (a higher variance estimator as noted in \sect\ref{sec:estimators}) relative to a true ATE of 0.4 for most tasks. We consider additional means of quantifying similarity, the mean ratio of Twitter:Reddit results, and Pearson's $r$ correlation. Mean absolute difference provides the clearest comparison, as ratios are strongly influenced by small but similar results (\eg Placebo Test). Pearson's $r$ miscommunicates the similarity of results when performance estimates cluster so closely that there is no consistent trend (but performance is highly similar between estimates and datasets). }
\label{tab:overall_reddit_twitter_similarity}
\end{table}

\new{
To compute overall similarity, we calculate the mean absolute difference between the results on the Twitter and Reddit versions of each task (Table~\ref{tab:overall_reddit_twitter_similarity}). We do this separately for each evaluation metric (\eg Treatment Accuracy and estimated $ATE$). When considering all five tasks, the mean absolute difference between Twitter and Reddit results is 0.019 across Treatment Accuracy for values between 0 and 1. This similarity holds when comparing $ATE$ estimates, with a mean absolute difference of 0.032 when comparing Stratified $ATE$ estimates, relative to a true $ATE$ of 0.4 for most tasks. As the IPTW estimator is more sensitive given the reweighting scheme and small probabilities (\sect\ref{sec:estimators}), the mean absolute difference between IPTW $ATE$ estimates is slightly larger, at 0.052. 
}

\begin{figure*}
    \centering
    \vspace{-12pt}
    \centerline{\includegraphics[height=.97\textheight]{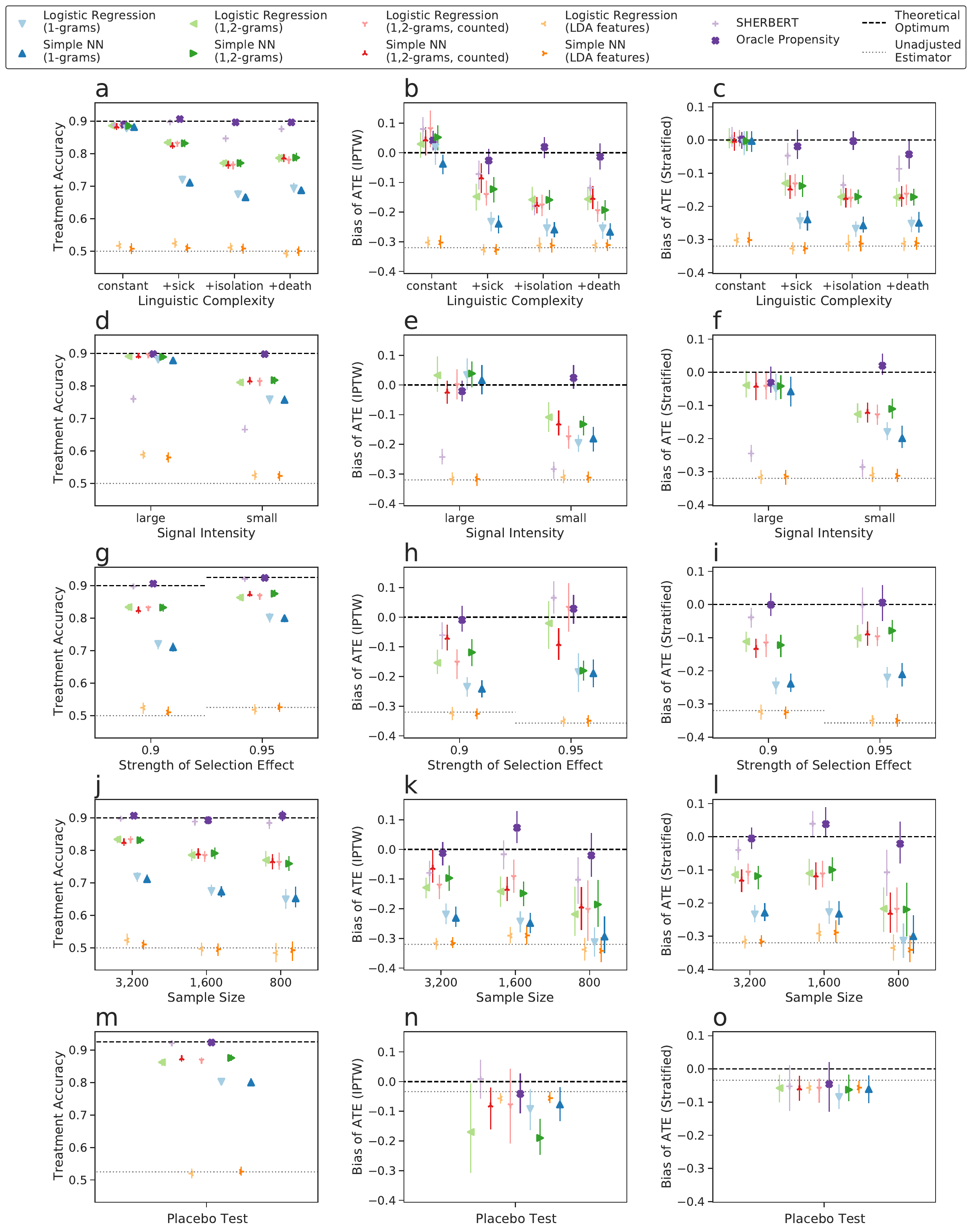}}
    \caption{\new{Results for tasks evaluated with Twitter data, with bootstrapped 95\% confidence intervals. Twitter results are extremely similar to those computed with Reddit data, with a mean absolute difference between Treatment Accuracies of 1.9 percentage points.}} 
    \label{fig:twitter_results}
\end{figure*}

\new{
We also compute additional similarity metrics, the mean ratio of Twitter:Reddit results, and the Pearson's correlation between Twitter and Reddit results (Table~\ref{tab:overall_reddit_twitter_similarity}). These metrics further emphasize the similarity between results computed on the two datasets, but are also strongly influenced by small but similar results, such as the true $ATE$ of 0 in the Placebo Test, in the case of the ratio between Twitter and Reddit, and by closely clustered results without consistent trends (\eg Figures~\ref{fig:twitter_results}o), in the case of Pearson's r.
}

\new{
Overall, we find that the model performance estimates are highly similar across both data sets with nearly identical performance statistics and relative trends across models and tasks.
}

\begin{figure*}
    \centering
    \vspace{-12pt}
    \centerline{\includegraphics[height=.97\textheight]{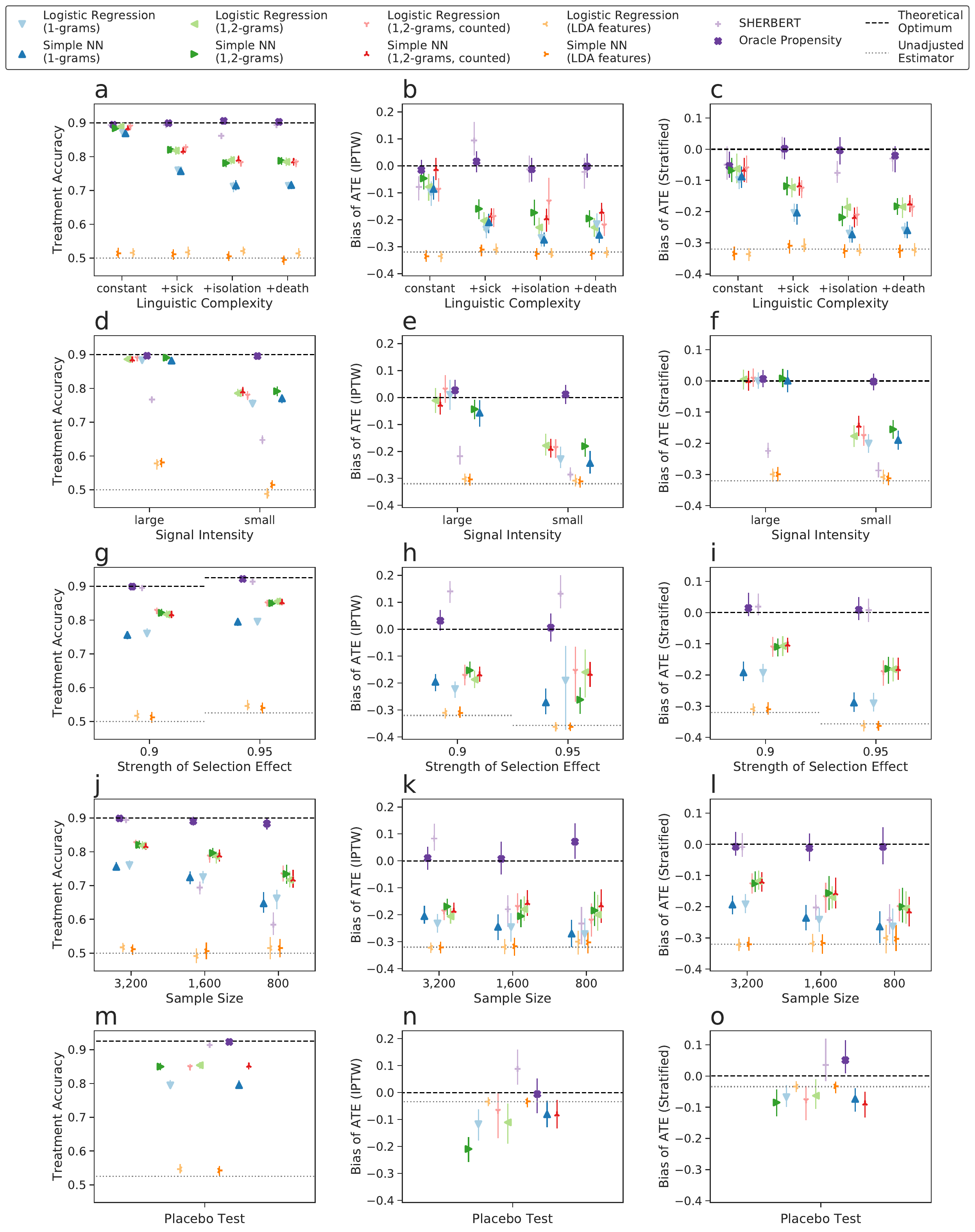}}
    \caption{\new{Results for tasks evaluated with Reddit data, repeated from Fig.~\ref{fig:results} for ease of comparison.}} 
    \label{fig:reddit_results_repeated}
\end{figure*}

\new{In this Appendix, on the following page, we first report complete Twitter results (Fig.~\ref{fig:twitter_results}), along with the Reddit results (Fig.~\ref{fig:reddit_results_repeated}) for ease of comparison. We then provide an additional details and analysis of the differences between Twitter and Reddit results (Appendix~\ref{appendix:reddit_twitter_differences}).}

\clearpage
\subsection{Comparison of Results across Twitter and Reddit Datasets}\label{appendix:reddit_twitter_differences}

\begin{figure*}[h]
    \centering
    \includegraphics[width=\textwidth]{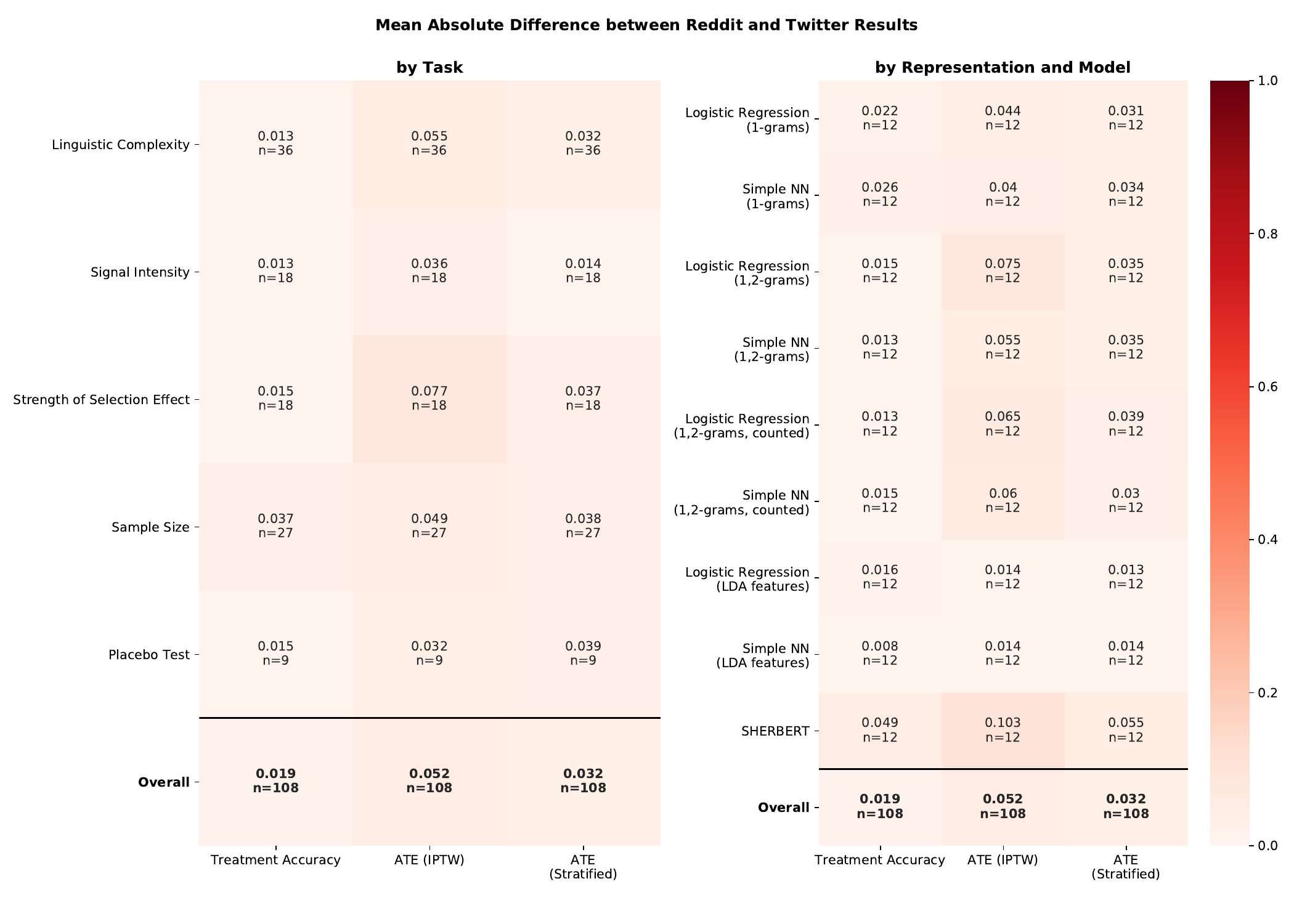}
    \caption{Comparison of Reddit and Twitter results, broken down by task and by text representation and model. Note that mean absolute differences are highly similar across the tasks and methods. Differences are slightly larger for $ATE$ estimates using IPTW, a higher variance estimator (as noted in \sect\ref{sec:estimators}), than for estimates using stratification. This further emphasizes the importance of evaluation metrics beyond accuracy.}
    \label{fig:twitter_comparison_correlations}
\end{figure*}

\new{
We compare correlations between Reddit and Twitter results separately for each task and for each text representation and propensity score model. Fig.~\ref{fig:twitter_comparison_correlations} displays the mean absolute difference between Reddit and Twitter results for each task (left) and text representation and model (right). The number of experiments used to compute the mean difference is given by n.
As with the overall results, prediction performance is highly similar across both datasets, consistently across tasks and models.
As expected, the increased sensitivity and variance of the IPTW $ATE$ estimate results in larger differences between Reddit and Twitter results than the more stable Stratified estimate (see \sect\ref{sec:estimators} for details).
}

\clearpage
\section{Order of Text Task}\label{appendix:order_of_text}

\begin{figure*}[h]
    \centering
    \centerline{\includegraphics[width=\textwidth]{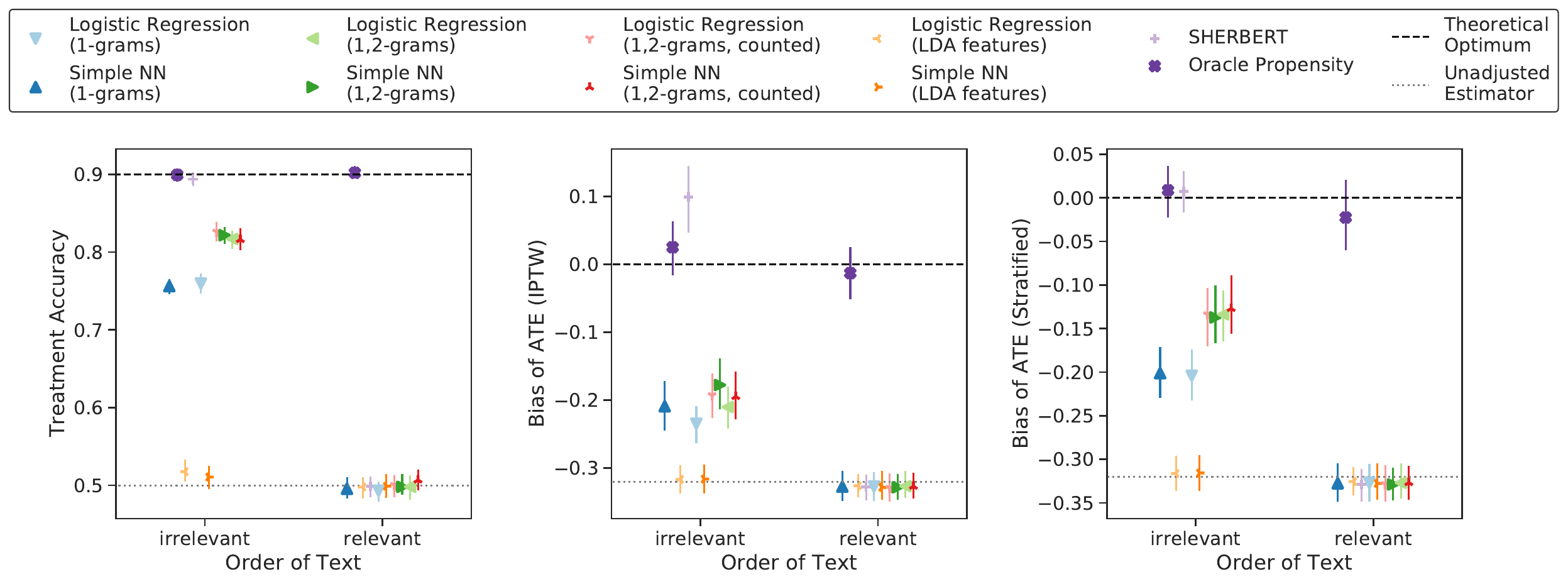}}
    \caption{Treatment Accuracy and Bias computed with IPTW and Stratification for the Order of Text task, \new{generated with Reddit data}.
    No model was able to distinguish between posts at the beginning and posts at the end of users' histories.} 
    \label{fig:order_of_text}
\end{figure*}

\new{
The order of social media posts can significantly influence their meaning and implications. For example, a person who posts about having good mental health and then posts a year later that they've lost a loved one might be at higher risk of developing suicidal ideation than a different person who first posts about losing a loved one and then immediately posts that their mental health is still good. Can models recognize order?
}

\new{
This task tests a model's ability to differentiate between the order of posts in a user's history.
There are two levels of difficulty.
For the baseline level, $f_1$ appends a random Sickness Post to the end of the user's history, and $f_2$ is simply the identity function.
At the harder level order becomes critical. Here , $f_1$ still appends a random Sickness Post to the end, but $f_2$ \textit{prepends} a random Sickness Post to the \textit{beginning} of a user's history.
The difference in performance between the two prediction tasks reflects the model's ability to differentiate temporal order.
}

\new{
The Order of Text task requires models to recognize models and reason about their order. The models tested incorporate information from long histories using simple aggregation across comments. Most aggregate by n-gram counting (including LDA, which compresses n-gram features), while \hbert uses a simple hierarchical attention network. Neither method captures this notion of order, so all models end up with effectively unadjusted estimator accuracy (Fig. \ref{fig:order_of_text}). Developing causal inference methods that capture these temporal dynamics is an important area of future work.
}

\end{document}